 \PassOptionsToPackage{table}{xcolor}
 
\documentclass[sigconf]{acmart}
\usepackage{amsmath,amsfonts,bm}

\def\eqref#1{equation~\ref{#1}}
\def\1{\bm{1}}

\DeclareMathAlphabet{\mathsfit}{\encodingdefault}{\sfdefault}{m}{sl}
\SetMathAlphabet{\mathsfit}{bold}{\encodingdefault}{\sfdefault}{bx}{n}

\usepackage{graphicx}
\usepackage{soul}
\usepackage{multirow}
\usepackage{booktabs}

\usepackage{enumitem}
\usepackage{balance}
\usepackage{bbding}
\usepackage{times}
\usepackage{url}
\usepackage{caption}
\usepackage{amsmath}
\usepackage{amsthm}
\usepackage{booktabs}
\usepackage{algorithm}
\usepackage{latexsym}

\usepackage{comment} 
\usepackage{amsmath}
\usepackage{amsfonts}
\usepackage{makecell}
\usepackage{color,soul}
\usepackage{multirow}
\usepackage{bm}
\usepackage{listings}
\usepackage{arydshln}
\usepackage{subfig}
\usepackage{algpseudocode}

\lstnewenvironment{python}[1][]
{\lstset{language=Python, basicstyle=\ttfamily\small, keywordstyle=\color{blue}, #1}}{}

\newcommand\Tstrut{\rule{0pt}{2.6ex}}       % "top" strut
\newcommand\Bstrut{\rule[-1.1ex]{0pt}{0pt}} % "bottom" strut
\usepackage{amsmath}
\definecolor{codegreen}{RGB}{0,185,18}
\definecolor{codegray}{rgb}{0.5,0.5,0.5}
\definecolor{codepurple}{rgb}{0.58,0,0.82}
\definecolor{backcolour}{RGB}{230, 230, 230}

\lstdefinestyle{mystyle}{
    backgroundcolor=\color{backcolour},   
    commentstyle=\color{codegreen},
    keywordstyle=\color{magenta},
    numberstyle=\tiny\color{codegray},
    stringstyle=\color{codepurple},
    basicstyle=\ttfamily\footnotesize,
    breakatwhitespace=false,         
    breaklines=true,                 
    captionpos=b,                    
    keepspaces=true,                 
    numbers=left,                    
    numbersep=5pt,                  
    showspaces=false,                
    showstringspaces=false,
    showtabs=false,                  
    tabsize=2
}

\newcommand{\SystemName}{OTJR}
\newcommand\mfont{\fontsize{7.6pt}{18pt}\selectfont}
\newcommand{\ImageNet}{I{\mfont MAGE}N{\mfont ET}}

\newcommand{\CIFAR}{C{\mfont IFAR}}

\newcommand{\WEB}{{\mfont WEB}}
\AtBeginDocument{%
  }

\setcopyright{none}\acmConference[arXiv]{}{2024}{USA}
\begin{document}

\title{Bridging Optimal Transport and Jacobian Regularization by Optimal Trajectory for Enhanced Adversarial Defense}


% \author{Anonymous submission}
\author{Binh M. Le}
\affiliation{%
  \institution{College of Computing and Informatics\\Sungkyunkwan University, S. Korea}
  \city{}
  \state{}
  \country{}
}
\email{bmle@g.skku.edu}
\author{Shahroz Tariq}
\affiliation{%
  \institution{
  Data61 CSIRO}
     \city{Sydney}
   \state{}
   \country{Australia}
}
\email{shahroz.tariq@csiro.au}
\author{Simon S. Woo}
\authornote{Corresponding author \Envelope}

\affiliation{%
  \institution{College of Computing and Informatics\\
  Sungkyunkwan University, S. Korea}
     \city{}
   \state{}
   \country{}
}
\email{swoo@g.skku.edu}
% \renewcommand{\shortauthors}{Submission number 225}
\renewcommand{\shortauthors}{Le et al.}
\begin{abstract}

% \simon{Add the motivation for the web...}

    % Deep neural networks are widely recognized as being vulnerable to adversarial perturbation. To overcome this challenge, developing a robust classifier is crucial.
    % So far, two well-known defenses have been adopted to improve the learning of robust classifiers, namely adversarial training (AT) and Jacobian regularization. However, each approach behaves differently against adversarial perturbations.
    % The Web, as a rich medium of diverse content, has been constantly under the threat of malicious entities exploiting its vulnerabilities, especially with the rapid proliferation of deep learning applications in various web services. One such vulnerability, crucial to the fidelity and integrity of web content, is the susceptibility of deep neural networks to adversarial perturbations, especially concerning images - a dominant form of data on the web. 
    Deep neural networks, particularly in vision tasks, are notably susceptible to adversarial perturbations. To overcome this challenge, developing a robust classifier is crucial. In light of the recent advancements in the robustness of classifiers, we delve deep into the intricacies of adversarial training and Jacobian regularization, two pivotal defenses.
    Our work is the first carefully analyzes and characterizes these two schools of approaches, both theoretically and empirically, to demonstrate how each approach impacts the robust learning of a classifier. Next, we propose our novel Optimal Transport with Jacobian regularization method, dubbed~\SystemName, bridging the input Jacobian regularization with the a output representation alignment by leveraging the optimal transport theory. In particular, we employ the Sliced Wasserstein distance that can efficiently push the adversarial samples' representations closer to those of clean samples, regardless of the number of classes within the dataset. The SW distance provides the adversarial samples' movement directions, which are much more informative and powerful for the Jacobian regularization.  Our empirical evaluations set a new standard in the domain, with our method achieving commendable accuracies of 52.57\% on ~\CIFAR-10 and 28.36\% on ~\CIFAR-100 datasets under the AutoAttack. Further validating our model's practicality, we conducted real-world tests by subjecting internet-sourced images to online adversarial attacks. These demonstrations highlight our model's capability to counteract sophisticated adversarial perturbations, affirming its significance and applicability in real-world scenarios.
    
    % Our extensive experiments demonstrate the effectiveness of our proposed method in defending well-known adversarial attacks, where we jointly incorporates Jacobian regularization into AT. Furthermore, we demonstrate that our proposed method consistently enhances the model's robustness over ~\CIFAR-10 and ~\CIFAR-100 datasets under various adversarial attack settings with \textit{four} different SOTA methods, achieving accuracy up to 51.41\% and 28.49\%, respectively,  under AutoAttack.
\end{abstract}

%    <concept>       <concept_id>10010147.10010257.10010258.10010261.10010276</concept_id>
%        <concept_desc>Computing methodologies~Adversarial learning</concept_desc>
%        <concept_significance>100</concept_significance>
%        </concept>
%  </ccs2012>

% \ccsdesc[100]{Computing methodologies~Adversarial Defense, Adversarial Learning, Adversarial Attacks}



\maketitle

\section{Introduction}
\begin{figure}[t!]
\centering
\includegraphics[width=1\linewidth]{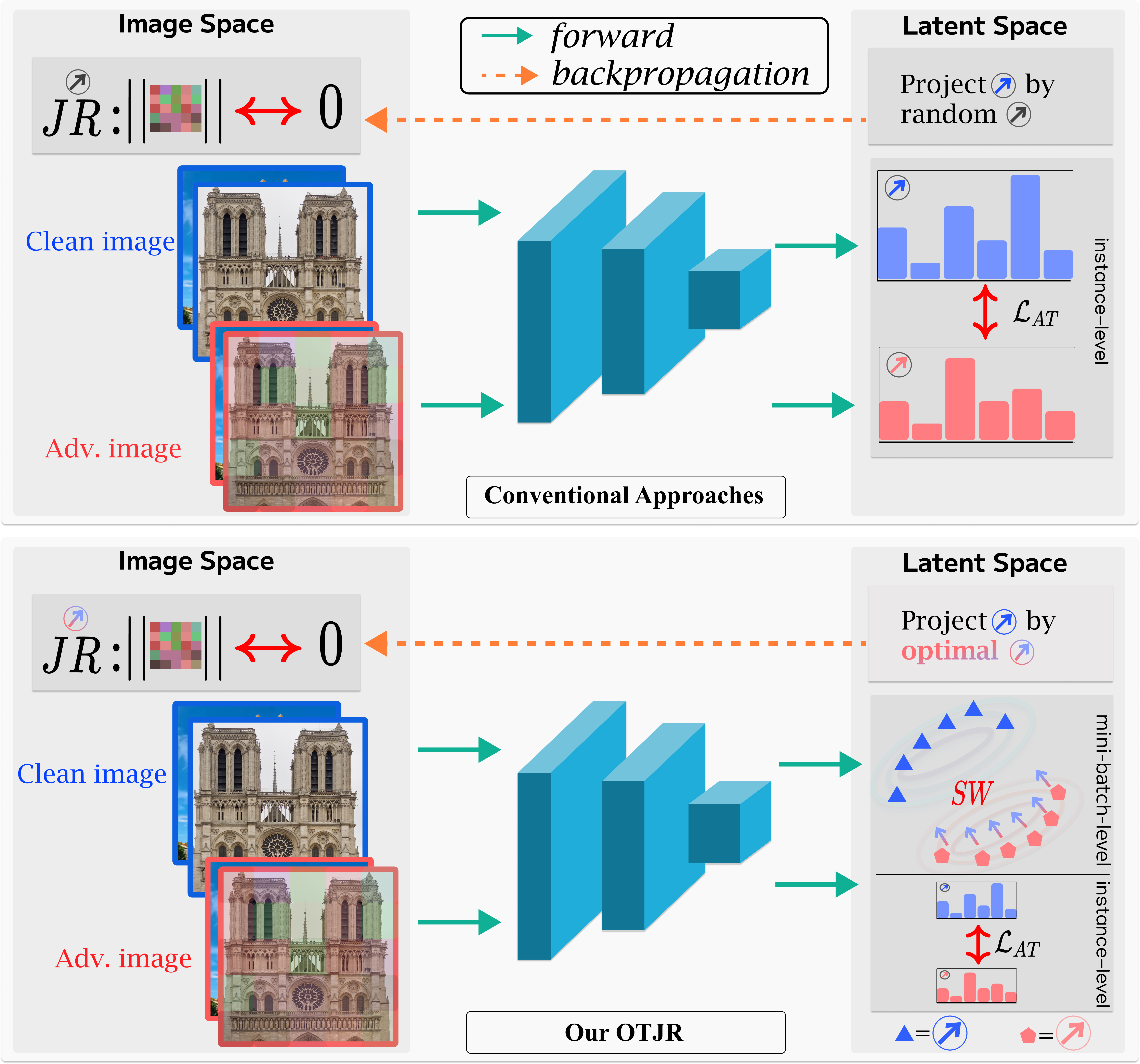}
\caption{    Illustration of (top) two popular approaches to boost a model's robustness: Adversarial Training (AT) vs. Jacobian regularization (JR), and (bottom) our OTJR method. Jacobian regularization  tries to silence the Jacobian matrix at the input end. The AT adjusts the distribution of perturbed samples at the output end. In conventional approach, Jacobian regularization backpropagates through random projections, whereas the AT via a loss function. Our proposed~\SystemName~ bridges AT and Jacobian regularization on framework by the optimal transport theory (Sliced Wasserstein distance).}
\label{fig:diff_impact}
\end{figure}

 Deep Neural Networks (DNNs) have established themselves as the de facto method for tackling challenging real-world machine learning problems.
 % The contemporary landscape of the Web, driven by dynamic services and rich content, heavily leans on Deep Neural Networks (DNNs) as the cornerstone for a myriad of complex machine learning challenges. 
Their  applications cover a broad range of domains, such as image classification, object detection, and recommendation systems.
Nevertheless, recent research has revealed DNNs' severe vulnerability to adversarial examples ~\cite{szegedy2013intriguing,goodfellow2014explaining}, particularly in computer vision tasks. Small imperceptible perturbations added to the image can easily deceive the neural networks into making incorrect predictions with high confidence. Moreover, this unanticipated phenomenon raises social concerns about DNNs' safety and trustworthiness, as they can be abused to attack many sophisticated and practical machine learning systems putting human lives into danger, such as in autonomous car~\cite{deng2020analysis} or medical systems ~\cite{ma2021understanding,bortsova2021adversarial}.

Meanwhile, there are numerous studies that devote their efforts to enhance the robustness of various models against adversarial examples. Among the existing defenses, adversarial training (AT) \cite{goodfellow2014explaining,madry2017towards} and Jacobian regularization (JR)~\cite{jakubovitz2018improving,hoffman2019robust} are the two most predominant and popular defense approaches \footnote{Another line of research investigates distributional robustness (DR) \cite{shafieezadeh2015distributionally, duchi2021statistics, gao2022wasserstein,rahimian2019distributionally, kuhn2019wasserstein,bui2022unified}, which seeks the worst-case distribution of generated perturbations (\textbf{generation step}). Notably, several studies have utilized the Wasserstein distance \cite{shafieezadeh2015distributionally,kuhn2019wasserstein,bui2022unified}. However, these investigations are beyond the scope of our current study (\textbf{optimization step}). On their own, they struggle to robustify a model and often necessitate integration with an AT loss. In our experimental section, we further demonstrate the superiority of a state-of-the-art DR method, namely UDR \cite{bui2022unified}, when combined with our approach.}.
In AT, small perturbations are added to a clean image in its neighbor of $L_p$ norm ball to generate adversarial samples. Thus, an adversarially trained model can force itself to focus more on the most relevant image's pixels. On the other hand, the second approach, Jacobian regularization, mitigates the effect of the perturbation to the model's decision boundary by suppressing its gradients. However, AT and Jacobian regularization have not been directly compared in both theoretical and empirical settings.

In this work, we embark on a dual-path exploration, offering both theoretical and empirical comparisons between AT and Jacobian regularization. Our objective is to deepen our comprehension of the adversarial robustness inherent to DNN models and subsequently enhance their defensive capability. While a myriad of prior research has spotlighted defense, they predominantly adopt either an empirical or theoretical lens, rarely both. To bridge this gap, we introduce an innovative approach, integrating both Jacobian regularization and AT. This fusion seeks to augment the adversarial robustness and defensive efficacy of a model, as elucidated in Fig. \ref{fig:diff_impact}.
 %In this work, we theoretically revisit how AT and Jacobian regularization have been differently derived for providing the adversarial robustness {and defensive effectiveness} of a model. Secondly, we conduct an extensive empirical analysis to demonstrate the distinct impacts of Jacobian regularization and AT on different layers of a defensive DNN.
% In addition, we observe that not only AT but also Jacobian regularization can effectively suppress the channel-wise activation achieving even better results. This remarkably counteracts what was stated by Bai \textit{et al.} \cite{bai2021improving} that adversarial robustness can be improved through suppressing the activation magnitude.
% Thus, we dismantle the bi-directional relationship: while increased adversarial robustness can implies smaller activation magnitudes, the inverse is not always true. \binh{This is just a observation, not a contribution}
% Furthermore, we demonstrate that AT and Jacobian regularization have different impacts on the model's behavior when perturbed samples are encountered. While AT frameworks support a model to adjust its learning parameters to become more stable under adversarial attacks, Jacobian regularization produces a more silent gradient map and thus directly reduces the perturbations' impact on the model's outputs. \binh{replicate the second paragraph}

For AT, a plethora of studies have been proposed, presenting unique strategies to encourage the learning of robust classifiers. \cite{zhang2020geometry,zhang2019theoretically,rony2019decoupling,kannan2018adversarial,madry2017towards}. Notably, Sinkhorn Adversarial Training (SAT) \cite{bouniot2021optimal} resonates with our methodology, particularly in its objective to bridge the distributional gap between clean and adversarial samples using optimal transport theory.
%in which the geometric distance between adversarial samples' and \hl{natural?} \simon{clean} samples' distributions is minimized using optimal transport theory. 
However, the pillar of their algorithms mainly relies on the Sinkhorn algorithm \cite{cuturi2013sinkhorn} to utilize the space discretization property \cite{vialard2019elementary}. Therefore, their approach has several limitations in terms of handling high-dimensional data~\cite{meng2021large,petrovich2020feature}. Particularly, the Sinkhorn algorithm blurs the transport plan by adding an entropic penalty to ensure the optimization’s convexity. The entropic penalty encourages the randomness of the transportation map. However, in high-dimensional spaces, such randomness reduces the deterministic movement plan of one sample, causing ambiguity.
% As a result, %FeaScatter and SAT results in a slow convergence rate when training defensive models on a large scale dataset, which is unjustifiable due to additional training epochs with distinct learning schedules are introduced \cite{pang2020bag}. 
As a result, when training defense models on a large
scale dataset, SAT results in a slow convergence rate, which is unjustifiable due to the introduction of additional training epochs with distinct learning schedules~\cite{pang2020bag}.

% To this end, we introduce the novel Optimal Transport with Jacobian regularization~\SystemName~method {for more effective defense against adversarial attacks}. 
To address the outlined challenges, we present our pioneering method, Optimal Transport with Jacobian Regularization, denoted as ~\SystemName, designed explicitly to bolster defenses against adversarial intrusions. We leverage the Sliced Wasserstein (SW) distance, which is more efficient for AT in high dimensional space with a faster convergence rate.
In addition, the SW distance provides us with other advantages due to optimal latent trajectories of adversarial samples in the embedding space, which is critical for designing an effective defense. We further integrate the input-output Jacobian regularization by substituting its random projections with the optimal trajectories and constructing the optimal Jacobian regularization. Our main contributions are summarized as follows: %footnote{Our code is available at \url{https://anonymous.4open.science/r/OTJR-2024}, and we plan to release the entire code upon acceptance of the paper.}:
% \begin{itemize}
    % \item \textbf{Comprehensive Theoretical and Empirical Examination} We first revisit the theory of adversarial robustness from the perspective of AT and Jacobian regularization. We are the first {to theoretically and empirically} analyze and provide a side-by-side comparison, and {lay the foundation for characterizing} the difference in impacts of each approach on a defensive DNN  design.
    
    % \item \textbf{Innovative Utilization of the Sliced Wasserstein (SW) Distance.} We propose the application of the SW distance in our AT, called  \SystemName, which can rapidly improve the convergence of the training procedure compared to prior works. With the support of SW, we derive \textbf{the optimal movement directions} of adversarial samples in the latent space. Then, we integrate these optimal directions into the Jacobian regularization, which can further increase the decision boundaries of DNNs.
    
    % \item \textbf{White- and Black-box Attack and Defense Performance.} Through extensive experiments, we show that our proposed method achieves higher performance compared to the well-known SOTA defense mechanisms, demonstrating the effectiveness and competitiveness of optimizing Jacobian regularization in AT as an effective defense mechanism. 

% \end{itemize}
\noindent \textbf{(1) Comprehensive Theoretical and Empirical Examination.} Our research delves deep into the theoretical underpinnings of adversarial robustness, emphasizing the intricacies of AT and Jacobian Regularization. Distinctively, we pioneer a simultaneous theoretical and empirical analysis, offering an incisive, comparative exploration. This endeavor serves to elucidate the differential impacts that each methodology has on the design and efficacy of defensive DNNs.

\noindent \textbf{(2) Innovative Utilization of the Sliced Wasserstein (SW) Distance.} Charting new territory, we introduce the integration of the SW distance within our AT paradigm, denoted as \SystemName. This innovation promises a marked acceleration in the training convergence, setting it apart from extant methodologies. Harnessing the prowess of the SW distance, we discern the optimal trajectories for adversarial samples within the latent space. Subsequently, we weave these optimal vectors into the framework of Jacobian Regularization, augmenting the resilience of DNNs by expanding their decision boundaries.

\noindent \textbf{(3) Rigorous Evaluation against White- and Black-box Attacks.} Our exhaustive experimental assessments underscore the superiority of our methodology. Pitted against renowned state-of-the-art defense strategies, our approach consistently emerges preeminent, underscoring the potency of enhancing Jacobian Regularization within the AT spectrum as a formidable defensive arsenal.

\section{Comparisons between AT and JR} %Jacobian Regularization}
\subsection{Theoretical Preliminaries}
Let a function $f$ represent a deep neural network (DNN), which is parameterized by $\theta$, and $x \in \mathbb{R}^{I}$ be a clean input image. Its corresponding output vector is $z=f(x) \in \mathbb{R}^C$, where $z_{c}$ is proportional to the likelihood that $x$ is in the
$c$-th class. Also, let $\tilde{x} = x + \epsilon$ be an adversarial sample of $x$ generated by adding a small perturbation vector $\epsilon \in \mathbb{R}^{I}$. Then, the Taylor expansion of the mapped feature of the adversarial sample with respect to $\epsilon$ is derived as follows:
\begin{align}
    \tilde{z} = f(x + \epsilon) &= f(x) + J(x) \epsilon + O (\epsilon ^{2}) \nonumber \\
    &\simeq z + J(x) \epsilon , 
\label{eqn:taylor}
\end{align}
\begin{equation}
    ||\tilde{z} - z||_{q} \simeq ||J(x) \epsilon ||_{q}.
\label{eqn:key_eqn}
\end{equation}
\begin{table}[t!]
\caption{  Training objectives of the popular AT frameworks, where $\mathbb{S}$ denotes the softmax function.}
\centering
\resizebox{0.96\linewidth}{!}{%
\begin{tabular}{l | c }
\Xhline{2\arrayrulewidth}
AT Framework            & Training Objective       \\
\hline \hline
TRADES    & $\mathcal{L}_{\mathcal{XE}}(\mathbb{S}(z), y) + \lambda \mathcal{L}_{\mathcal{XE}}(\mathbb{S}(\tilde{z}), \mathbb{S}(z))$    \Tstrut\\
% MART    & $\mathcal{L}_{\mathcal{XE}}(\mathbb{S}(z), y) + \lambda \mathcal{L}_{\mathcal{XE}}(\mathbb{S}(\tilde{z}), \mathbb{S}(z))\cdot (1-\mathbb{S}_{y}(z))$    \Tstrut\\
PGD-AT    &   $ \mathcal{L}_{\mathcal{XE}}(\mathbb{S}(\tilde{z}), y)$         \Tstrut\\
ALP  & $\alpha \mathcal{L}_{\mathcal{XE}}(\mathbb{S}(\tilde{z}), y) +(1-\alpha) \mathcal{L}_{\mathcal{XE}}(\mathbb{S}(\tilde{z}), y) + \lambda ||\tilde{z}-z||
_2$ \Bstrut\Tstrut\\
\Xhline{2\arrayrulewidth}
\end{tabular}%
}
\label{tb:stategies}
\end{table}
\begin{figure*}[!ht]
  \centering
\includegraphics[width=0.96\linewidth]{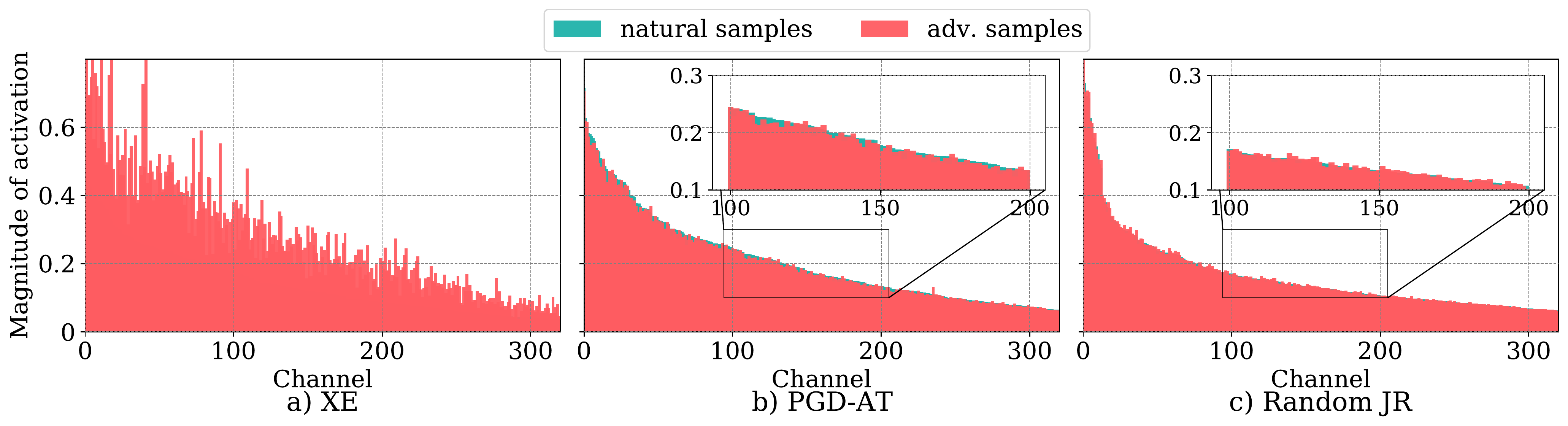}
\caption{  The magnitude of activation at the penultimate layer for models trained with $\mathcal{XE}$ loss, PGD-AT adversarial training, and the input-output Jacobian regularization. The channels in the X-axis are sorted in descending order of the clean samples' magnitude. }%
\label{fig:mag_activation}
\end{figure*}

Precisely, Eq.~\ref{eqn:key_eqn} is a basis to derive two primary schools of approaches, AT vs. Jacobian regularization, for mitigating adversarial perturbations and boosting the model robustness. In particular, each side of ~Eq.~\ref{eqn:key_eqn} targets the following two objectives, to improve the robustness: 

    \textbf{\textit{1) Aligning adversarial representation (AT)}.} Minimizing the left-hand side of Eq.~\ref{eqn:key_eqn} is to push the likelihood of an adversarial sample $\tilde{x}$ close to that of a clean sample $x$. For instance, the Kullback-Leibler divergence between two likelihoods is a popular AT framework such as TRADES~\cite{zhang2019theoretically}. More  broadly, the likelihood differences can include the cross-entropy ($\mathcal{XE}$) loss of adversarial samples such as ALP \cite{kannan2018adversarial}, PGD-AT \cite{madry2017towards}, or FreeAT \cite{shafahi2019adversarial}. These well-known AT frameworks are summarized in Table~\ref{tb:stategies}, to explain their core learning objective.

    \textbf{\textit{2) Regularizing input-output Jacobian matrix (JR)}.} Regularizing the right-hand side of Eq.~\ref{eqn:key_eqn}, which is independent of $\tilde{x}$, suppresses the spectrum of the input-output Jacobian matrix $J(x)$. Thus, the model becomes more stable with respect to input perturbation, as it was theoretically and empirically demonstrated in a line of recent research \cite{jakubovitz2018improving,hoffman2019robust,co2021jacobian}. Particularly, by observing $||J(x) \epsilon ||_{q} \leq ||J(x)||_F ||\epsilon || $, one can instead minimize the square of the Frobenius norm of the Jacobian matrix, which can be estimated as follows \cite{hoffman2019robust}:
    \begin{equation}
        ||J(x)||_{F}^{2} = C\mathbb{E}_{\hat{v}\sim\mathcal{S}^{C-1}} \left[ ||\hat{v} \cdot J ||^2\right],
    \label{eqn:estimator}
    \end{equation}
    where $\hat{v}$ is an uniform random vector drawn from a \textit{C}-dimensional unit sphere $\mathcal{S}^{C-1}$. Using Monte Carlo method to approximate the integration of $\hat{v}$ over the unit sphere,  Eq.~\ref{eqn:estimator} can be rewritten as:
    \begin{equation}
        ||J(x)||_{F}^{2} = \frac{1}{n_\text{proj}} \sum_{i=1}^{n_\text{proj}} \left[ \frac{\partial (\hat{v}_i \cdot z)}{\partial x}\right]^{2}.
    \end{equation}
    Considering a large number of samples in a mini-batch, $n_\text{proj}$ is usually set to $1$ for the efficient computation. Hence, the Jacobian regularization is expressed as follows:
    \begin{equation}
    \label{eqn:one_jac}
    ||J(x)||_{F}^{2} \simeq \left[ \frac{\partial (\hat{v} \cdot z)}{\partial x}\right]^{2},  \hat{v} \sim \mathcal{S}^{C-1}.
    \end{equation}

\noindent \textbf{Summary.} As shown above, each approach takes different direction to tackle adversarial perturbations.
%follow different directions.
While AT targets aligning the likelihoods at the output end, Jacobian regularization forces the norm of the Jacobian matrix at the input to zero, as also pictorially described in Fig.~\ref{fig:diff_impact}. However, their distinct effects on the robustness of a DNN have not been fully compared and analyzed so far.
\subsection{Empirical Analysis}
% \begin{figure*}[ht!]

%     \centering
%     \includegraphics[width=12.0cm]{  Figures/main_magnitude_activation.png}
%     % \centering
%     % \subfloat[\centering \small $\mathcal{XE}$ loss]{{\includegraphics[width=3cm]{ Figures/magnitude_activateion_xe_pgd25_nolegend.pdf} }}\hfill
%     % \subfloat[\centering \small PGD-AT]{{\includegraphics[width=3cm]{ Figures/magnitude_activateion_pgd_pgd25.pdf} }}\hfill
%     % \subfloat[\centering \small JR ]{{\includegraphics[width=3cm]{ Figures/magnitude_activateion_rand_jac_pgd25.pdf} }}\hfill
%     % \subfloat{{\includegraphics[width=2.0cm]{ Figures/legend_fig2.pdf} }}\hfill
%     \caption{   The magnitude of activation at the penultimate layer for models trained with $\mathcal{XE}$ loss, PGD-AT adversarial training, and the input-output {Jacobian} regularization. The channels in the X-axis are sorted in descending order of the clean samples' magnitude. More visualizations are provided in \textit{Supp. Material}. \sh{ I think it will save space if we combine figure 2 and 3 like I did above.}}%
%     \label{fig:mag_activation}
% \end{figure*}
% \label{subsec:exp_analysis}
% \begin{figure}[t!]
% \centering
% \includegraphics[width=6.9cm]
% { Figures/Jac_vs_PGD_input_der.pdf}
% \caption{   The magnitude of $||\nabla_{x} \mathcal{L}_{\mathcal{XE}}||_1$ at the input layer for a model trained with PGD-AT and Jacobian regularization, respectively. The red and green-filled areas range from minimum to maximum values of each sample.}
% \label{fig:input_der}
% \end{figure}

We also conduct an experimental analysis to ascertain and characterize the distinct effects of the two approaches (AT vs. Jacobian regularization) on a defense DNN when it is trained with each of the objectives. We use wide residual network WRN34 as the preliminary baseline architecture on~\CIFAR-10 dataset and apply two canonical frameworks: PGD-AT \cite{madry2017towards} and input-output Jacobian regularization \cite{hoffman2019robust} to enhance the model's robustness. Details of training settings are provided in the experiment section.

\begin{figure}[t!]
 \centering
 \includegraphics[width=8.2cm]{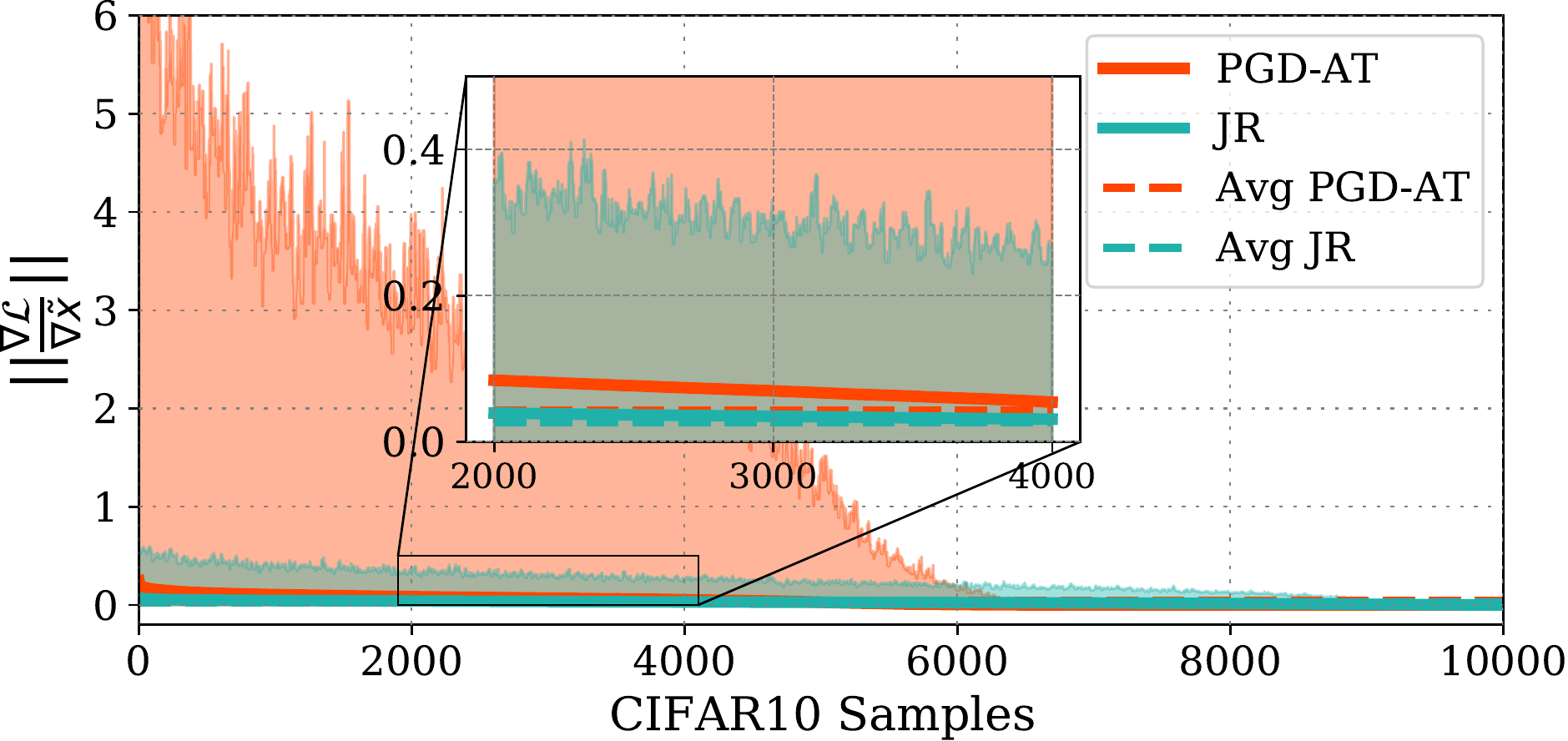}
 \caption{  Magnitude of $||\nabla_{\tilde{x}} \mathcal{L}_{\mathcal{XE}}||_1$ at the input layer for a model trained with PGD-AT and Jacobian regularization. Red and green-filled areas range from min. to max. values of each sample.}
 \label{fig:input_der}
\end{figure}

\noindent \textbf{Output and input sides.} We measure the magnitude of channel-wise activation to characterize the connection between adversarial defense methods and the \textit{penultimate layer's activation} \cite{bai2021improving}. Figure~\ref{fig:mag_activation} provides the average magnitude of activations of clean vs. adversarial samples created by PGD-20 attacks \cite{madry2017towards}. As shown, not only AT~\cite{bai2021improving}, but also Jacobian regularization can effectively suppress the magnitude of the activation. Moreover, the Jacobian regularization typically achieves the lower magnitude value of the activation compared to that of PGD-AT. This observation serves as a clear counter-example to earlier results from ~\citeauthor{bai2021improving}\cite{bai2021improving}, where they claim that adversarial robustness can be generally achieved via channel-wise activation suppressing. As such, it is worth noting that while a more effective defense strategy can produce lower activation, the inverse is not always true. In addition, Fig.~\ref{fig:input_der} represents the average gradient of $\mathcal{XE}$ loss with respect to the adversarial samples, at the \textit{input layer}. This demonstrates that the model trained with Jacobian regularization suppresses input gradients more effectively than a typical AT framework, \textit{i.e.}, PGD-AT. 
In other words, when a defensive model is abused to generate adversarial samples, pre-training with Jacobian regularization can reduce the severity of perturbations, hindering the adversary's target.
\begin{figure}[t!]
\centering
\includegraphics[width=8.4cm]{ 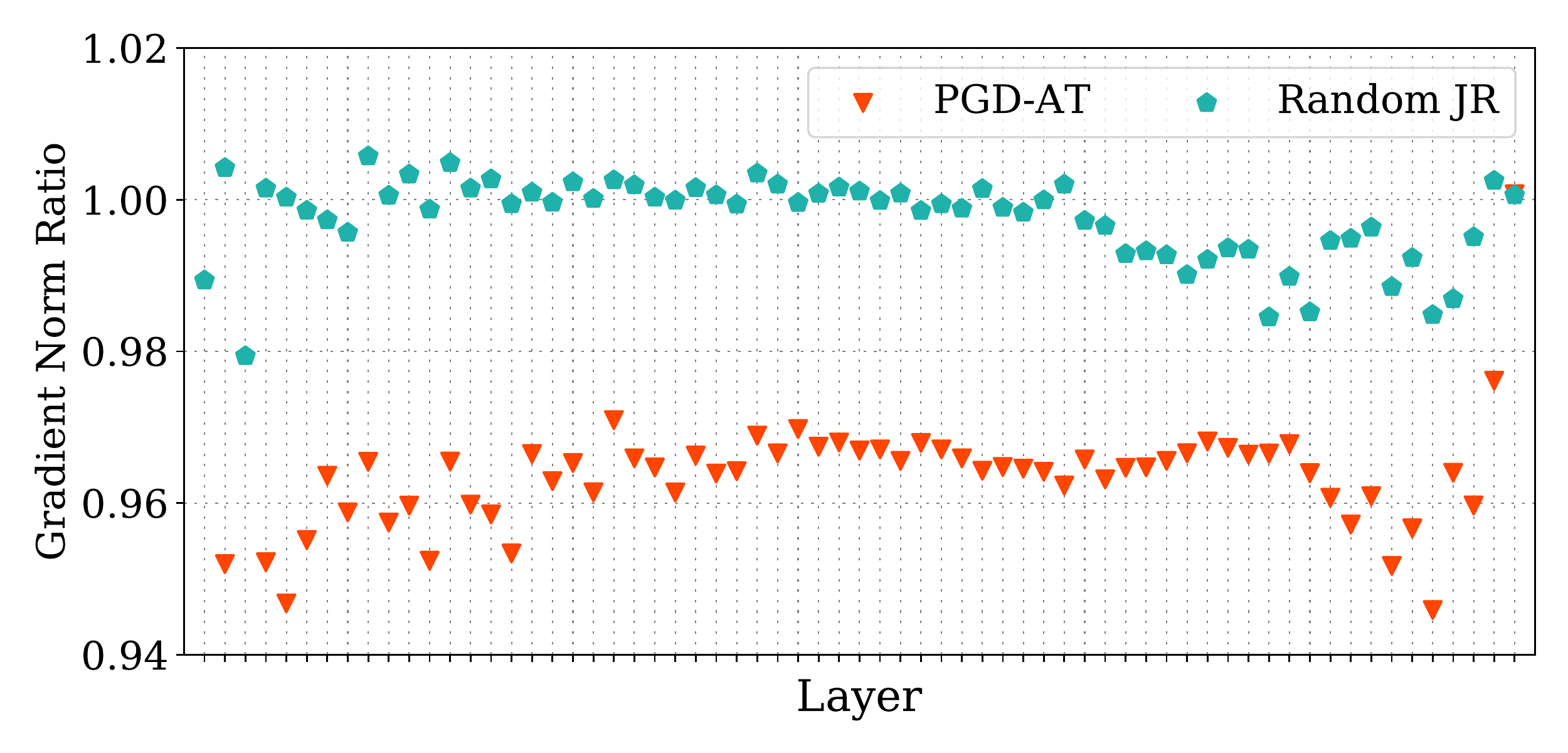}
\caption{   Ratios of $\mathbb{E}(||\nabla_{\theta_i}\mathcal{L}(\tilde{x})|| / ||\nabla_{\theta_i}\mathcal{L}({x})||)$ {w.r.t.} the model's parameters $\theta_{i}$ on~\CIFAR-10. The lower the ratios are, the more emphasis the model puts on perturbations.}
\label{fig:model_der}
\end{figure}

\noindent \textbf{Layer-by-layer basis}. We further provide Fig.~\ref{fig:model_der} {to} depict the gradients of models trained with PGD-AT, and Jacobin regularization, respectively. Particularly, we compute the norm ratios of the loss gradient on the adversarial sample to the loss gradient on the clean samples for each layer of the model. As we can observe, the model trained with the Jacobian regularization produces higher ratio values, meaning it puts less emphasis on adversarial samples due to its agnostic defense mechanism. Meanwhile,  most of the ratio values at the middle layers from Jacobian training vary around 1.This is explained by the regularization applied to its first derivatives. In summary, we can also empirically conclude that the Jacobian regularization tends to silence the gradient of the model from output to input layers. Therefore, it \textit{agnostically} stabilizes the model under the changes of input samples, and produces low-magnitude adversarial perturbations, when the model is attacked. In contrast, by learning the meaningful pixels from input images, AT adjusts the model's parameters at every layer in such a way to reduce the impacts of adversarial perturbation on the model's outputs. 

\noindent \textbf{Our motivation}. As elucidated  by~\citeauthor{hoffman2019robust}\cite{hoffman2019robust},  the computational overhead introduced by training models with Jacobian regularization is marginal compared to standard training regimes. Therefore, a combination of AT and Jacobian regularization becomes an appealing approach for the adversarial robustness of a model. Furthermore, taking advantages from both approaches can effectively render a classifier to suppress the perturbation and adaptively learn crucial features from both clean and adversarial samples. However, merely adding both approaches together into the training loss is not the best option. Indeed it is insufficient, since the adversarial representations in the latent space can contain meaningful information for the Jacobian regularization, which we will discuss more in the next section.

%However, it is trivial and insufficient to barely add them together into the training loss since the adversarial representations in the latent space can embrace meaningful information for the Jacobian regularization. 

Hence, in this work, we propose a novel optimization framework,~\SystemName, to leverage the movement direction information of adversarial samples in the latent space and optimize the Jacobian regularization. In this fashion, we can successfully establish a relationship and balance between silencing input's gradients and aligning output distributions, and significantly improve the overall model's robustness.
Additionally, recent studies proposed approaches utilizing a surrogate model \cite{wu2020adversarial} or teacher-student framework \cite{cui2021learnable} during training. Yet, while these methods improve the model's robustness, they also rely on previous training losses (such as TRADE or PGD). And, they introduce additional computation for the AT, which are so far well-known for their slow training speed and computational overhead. In our experiment, we show that our novel training loss can be compatible with these frameworks and further improve the model's robustness by a significant margin compared to prior losses.

\section{Our Approach}
Our approach explores the Sliced Wasserstein  distance in order to push the adversarial distribution closer to the natural distribution with a faster convergence rate. Next, \textit{\textbf{the Sliced Wasserstein distance results in optimal movement directions to sufficiently minimize the spectrum of input-output Jacobian matrix.}}

\begin{figure}[!b]
    \centering
    \includegraphics[width=0.42\textwidth]{ 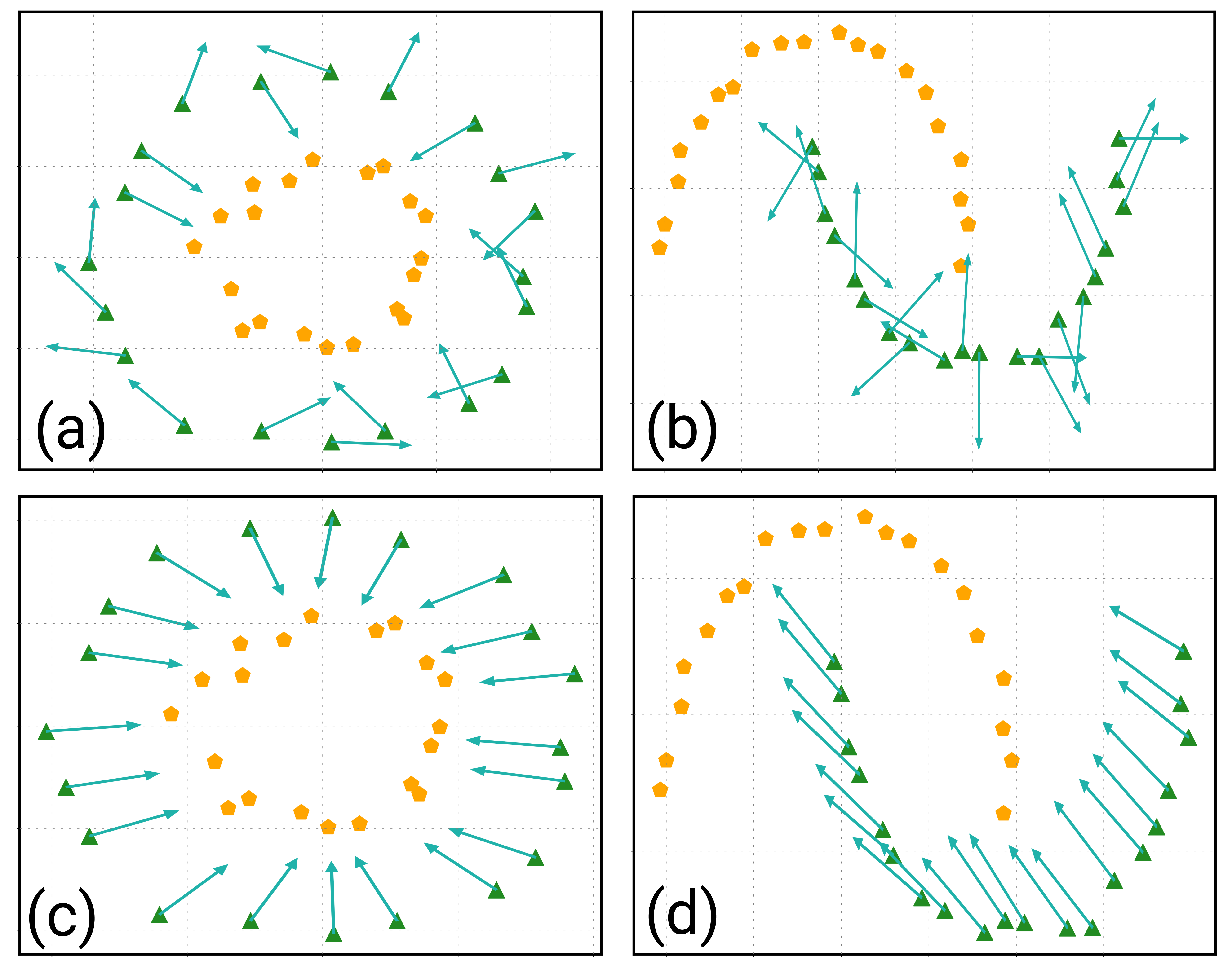}
    \caption{ Illustration of optimal latent trajectories. \textbf{Top row} ((a) \& (b)): Random movement directions  (green arrows), which is non-informative, uniformly sampled from two-dimensional unit sphere $\mathcal{S}^{1}$. \textbf{Bottom row} ((c) \& (d)): The optimal trajectories from the SW distance between source distribution (\textcolor{green!65!black}{green}) and target distribution (\textcolor{orange}{orange}) obtained from ~Eq.~\ref{eqn:opt_move}.}
    \label{fig:opt_move_toy}
\end{figure}

\begin{table*}[!t]
\centering
\caption{\textbf{Classification accuracy (\% ) under \textit{white-box}, \textit{black-box} attacks and \textit{AutoAttack}.} Different defense methods trained on~\CIFAR-10 and~\CIFAR-100 datasets using WRN34 in 100 epochs, except that SAT$^{400}$ was trained in 400 epochs.  Each experiment was conducted three times to ensure reliability. Standard deviations are denoted with subscripts. \underline{Numbers underlined} represent the greatest reduction in performance on clean images to trade-off for robustness. Best performances are highlighted in bold. }
\resizebox{0.95\textwidth}{!}{%
    \begin{tabular}{c | l | c | c c c c c c c | c c | c } 
    \Xhline{3\arrayrulewidth}
    Dataset & Defense    &\textit{Clean}    &PGD$^{20}$    &PGD$^{100}$    &$L_{2}$-PGD       &MIM &FGSM   &CW     &FAB   &Square & SimBa  &\emph{AutoAtt}\Tstrut\Bstrut\\
    \Xhline{2\arrayrulewidth}
    \multirow{8}{*}{\rotatebox[origin=c]{90}{\centering \small{~\CIFAR-10}}}    
&TRADES & \textit{84.71}$_{.15}$ & 54.23$_{.07}$ & 53.91$_{.11}$ & 61.04$_{.21}$ & 54.13$_{.11}$ & 60.46$_{.15}$ & 53.17$_{.22}$ & 53.65$_{.39}$ & 62.25$_{.26}$ & 70.66$_{.52}$ & 52.06$_{.15}$ \Tstrut\\
&ALP & \textit{86.63}$_{.23}$ & 46.99$_{.25}$ & 46.48$_{.25}$ & 55.69$_{.33}$ & 46.88$_{.23}$ & 58.14$_{.21}$ & 47.50$_{.30}$ & \textbf{55.61}$_{.27}$ & 59.26$_{.28}$ & 68.28$_{.31}$ & 46.28$_{.25}$ \\
&PGD-AT & \textit{86.54}$_{.20}$ & 46.67$_{.11}$ & 46.23$_{.12}$ & 55.93$_{.10}$ & 46.66$_{.09}$ & 56.83$_{.23}$ & 47.56$_{.15}$ & 48.39$_{.22}$ & 58.92$_{.12}$ & 68.66$_{.14}$ & 45.77$_{.07}$ \\
&SAT & \textit{\underline{83.19}}$_{.33}$ & 53.52$_{.15}$ & 53.23$_{.23}$ & 60.37$_{.06}$ & 52.46$_{1.34}$ & 60.36$_{.13}$ & 52.15$_{.20}$ & 52.54$_{.36}$ & 61.50$_{.19}$ & 69.70$_{.29}$ & 50.73$_{.20}$ \Bstrut\\
    \cline{2-13}
    &\textit{Random JR}   & \textit{84.99}$_{.14}$ & 22.67$_{.15}$ & 21.89$_{.14}$ & 60.98$_{.19}$ & 22.49$_{.18}$ & 32.99$_{.21}$ & 22.00$_{.06}$ & 21.74$_{.11}$ & 45.86$_{.20}$ & 71.20$_{.31}$ & 20.54$_{.14}$ \Tstrut\\
&\textit{SW}  &  \textit{84.26}$_{.80}$ & 54.51$_{.33}$ & 54.20$_{.37}$ & 61.08$_{.12}$ & 54.46$_{.27}$ & 61.32$_{.12}$ & 53.78$_{.05}$ & 54.90$_{.29}$ & 62.95$_{.43}$ & 70.74$_{.65}$ & 51.97$_{.06}$ \\
    &\cellcolor{backcolour}\SystemName~(\emph{\textbf{ours}}) & \cellcolor{backcolour}\textit{84.01}$_{.53}$ & \cellcolor{backcolour}\textbf{55.38}$_{.29}$ & \cellcolor{backcolour}\textbf{55.08}$_{.36}$ & \cellcolor{backcolour}\textbf{63.87}$_{.09}$ & \cellcolor{backcolour}\textbf{55.31}$_{.29}$ & \cellcolor{backcolour}\textbf{61.03}$_{.18}$ & \cellcolor{backcolour}\textbf{54.09}$_{.12}$ & \cellcolor{backcolour}{54.17}$_{.07}$ & \cellcolor{backcolour}\textbf{63.11}$_{.21}$ & \cellcolor{backcolour}\textbf{72.04}$_{.68}$ & \cellcolor{backcolour}\textbf{52.57}$_{.12}$ \Bstrut \\
    \Xhline{3\arrayrulewidth}
    \multirow{8}{*}{\rotatebox[origin=c]{90}{\centering \small{~\CIFAR-100}}}    
    &TRADES   & 57.46$_{.16}$ & 30.42$_{.05}$ & 30.36$_{.09}$ & 35.85$_{.10}$ & 30.37$_{.08}$ & 33.04$_{.16}$ & 27.97$_{.13}$ & 27.93$_{.15}$ & 33.70$_{.06}$ & 44.28$_{.14}$ & 27.15$_{.09}$\Tstrut\\
&ALP    & \textit{60.61}$_{.05}$ & 26.23$_{.18}$ & 25.87$_{.17}$ & 33.75$_{.09}$ & 26.14$_{.07}$ & 31.40$_{.08}$ & 25.78$_{.22}$ & 25.69$_{.15}$ & 33.09$_{.15}$ & 43.25$_{.22}$ & 24.57$_{.22}$ \\
&PGD-AT &  \textit{59.7}7$_{.21}$ & 24.05$_{.03}$ & 23.74$_{.03}$ & 31.41$_{.16}$ & 24.01$_{.06}$ & 29.21$_{.18}$ & 24.67$_{.04}$ & 24.47$_{.06}$ & 31.47$_{.11}$ & 41.15$_{.21}$ & 23.28$_{.01}$ \\
&SAT$^{400}$    & \textit{\underline{53.61}}$_{.52}$ & 26.63$_{.14}$ & 26.42$_{.16}$ & 32.34$_{.25}$ & 26.57$_{.19}$ & 31.04$_{.21}$ & 25.22$_{.06}$ & 26.63$_{.11}$ & 31.32$_{.34}$ & 39.64$_{.90}$ & 24.32$_{.05}$ \Bstrut\\ 
    \cline{2-13}
    &\textit{Random JR}   & \textit{66.58}$_{.17}$ & 9.41$_{.44}$ & 8.87$_{.42}$ & 37.79$_{.31}$ & 9.27$_{.49}$ & 16.38$_{.33}$ & 10.27$_{.45}$ & 9.26$_{.23}$ & 23.53$_{.15}$ & 48.86$_{.55}$ & 8.10$_{.57}$ \Tstrut\\
&\textit{SW}     & \textit{57.69}$_{.28}$ & 26.01$_{.16}$ & 25.82$_{.24}$ & 31.37$_{.17}$ & 25.97$_{.18}$ & 30.78$_{.34}$ & 25.48$_{.23}$ & 26.11$_{.07}$ & 31.34$_{.26}$ & 40.68$_{.43}$ & 24.35$_{.29}$ \\
    &\cellcolor{backcolour}\SystemName~(\emph{\textbf{ours}}) & \cellcolor{backcolour}\textit{58.20}$_{.13}$ & \cellcolor{backcolour}\textbf{32.11}$_{.21}$ & \cellcolor{backcolour}\textbf{32.01}$_{.18}$ & \cellcolor{backcolour}\textbf{43.13}$_{.12}$ & \cellcolor{backcolour}\textbf{32.07}$_{.19}$ & \cellcolor{backcolour}\textbf{34.26}$_{.30}$ & \cellcolor{backcolour}\textbf{29.71}$_{.06}$ & \cellcolor{backcolour}\textbf{29.24}$_{.08}$ & \cellcolor{backcolour}\textbf{36.27}$_{.05}$ & \cellcolor{backcolour}\textbf{49.92}$_{.23}$ & \cellcolor{backcolour}\textbf{28.36}$_{.10}$ \Bstrut \\
    \Xhline{3\arrayrulewidth}
    \end{tabular}%
}
\label{tb:white_attack}
\end{table*}

\subsection{Sliced Wasserstein Distance} The \emph{p}-Wasserstein distance between two probability distributions $\mu$ and $\nu$ \cite{villani2008optimal} in a general \emph{d}-dimensional space $\Omega$, to search for an optimal transportation cost between $\mu$ and $\nu$, is defined as follows:
\begin{equation}
\label{eqn:p-wasseitern}
\small
    W_p(P_{\mu},P_{\nu}) = \Big(\inf_{\pi \in \Pi(\mu, \nu)}  \int_{\Omega \times \Omega}\psi(x,y)^p d\pi(x,y) \Big)^{1/p},
\end{equation}
where $\psi: \Omega \times \Omega \rightarrow \mathbb{R}^{+}$ is a transportation cost function, and $\Pi(\mu, \nu)$ is a collection of all possible transportation plans. The Sliced \emph{p}-Wasserstein distance ($SW_p$), which is inspired by the $W_p$ in one-dimensional, calculates the \emph{p}-Wasserstein distance by projecting $\mu$ and $\nu$ onto multiple one-dimensional marginal distributions using Radon transform \cite{helgason2010integral}. The $SW_p$ is defined as follows:
\begin{equation}
\small
    SW_{p}(\mu, \nu) = \int_{\mathcal{S}^{d-1}}W_{p}(\mathcal{R}_{(t,\hat{v})}\mu, \mathcal{R}_{(t,\hat{v})}\nu) d\hat{v},
\end{equation}
where $\mathcal{R}_{(t,\hat{v})}\mu$ is the Radon transform as follows:
\begin{equation}
\small
    \mathcal{R}_{(t,\hat{v})}\mu = \int_{\Omega} \mu (x) \sigma (t-\langle \hat{v}, x \rangle) dx, \forall \hat{v} \in \mathcal{S}^{d-1}, \forall t \in \mathbb{R},
\end{equation}
% \begin{figure}{r}
%     \centering
%     \includegraphics[width=0.38\textwidth]{ Figures/optimal_movement.png}
%     \caption{ Illustration of optimal movement directions. \textbf{Top row} ((a) \& (b)): Random movement directions  (green arrows), which is non-informative, uniformly sampled from two-dimensional unit sphere $\mathcal{S}^{1}$. \textbf{Bottom row} ((c) \& (d)): The optimal movement directions from the SW distance between source distribution (\textcolor{green!65!black}{green}) and target distribution (\textcolor{orange}{orange}) obtained from ~Eq.~\ref{eqn:opt_move}.}
%     \label{fig:opt_move_toy}
% \end{figure}
where $\langle \cdot, \cdot \rangle$ denote the Euclidean inner product, and $\sigma$ is the Dirac delta function.

Next, let $\mathcal{B}$ denote the size of a mini-batch of samples, and $C$ denote the number of classes. To calculate the transportation cost between adversarial samples' representations $\mu :  = \{\tilde{z}_{i}\}_{\mathcal{B}}$ and the corresponding original samples' representations $\nu :  = \{z_i\}_{\mathcal{B}}$, the integration of $\hat{v}$ over the unit sphere $\mathcal{S}^{C-1}$ is approximated via Monte Carlo method with $K$ uniformly sampled random vector $\hat{v}_i \in \mathcal{S}^{C-1}$. In particular, $\mathcal{R}_{(t,\hat{v})}\mu$ and $\mathcal{R}_{(t,\hat{v})}\nu$ are sorted in ascending order using two permutation operators $\tau_1$  and $\tau_2$, respectively, and the approximation of Sliced \textit{1}-Wasserstein is expressed as follows:
\begin{equation}
\label{eqn:swd_approx}
\small
    SW(\mu, \nu) \simeq  \sum_{k=1}^{K}  \psi  \left(\tau_1 \circ \mathcal{R}_{(t,\hat{v}_k)}\mu, \tau_2 \circ\mathcal{R}_{(t,\hat{v}_k)}\nu \right).
\end{equation}

 \subsection{Optimal Latent Trajectory} Using~Eq.~\ref{eqn:swd_approx}, we can straightforwardly compute the trajectory of each $\tilde{z}_{i}$ in the latent space $\mathbb{R}^{C}$ in order to minimize the $SW(\mu, \nu)$. We refer to the directions of these movements as \textit{optimal latent trajectories}, because the SW distance produces the lowest transportation cost between the source and target distributions. Particularly, let the trajectories of $\{\tilde{z}_i \}_{\mathcal{B}}$ in each single projection $\hat{v}_k$ be:
\begin{equation}
    m_k = (\tau_1^{-1} \circ \tau_2 \circ \mathcal{R}_{(t,\hat{v}_k)}\nu - \mathcal{R}_{(t,\hat{v}_k)}\mu ) \otimes \hat{v}_k,
\end{equation}
where $\otimes$ denotes the outer product. Then, the overall optimal trajectory direction of each $\tilde{z}_{i}$ is expressed as follows:
\begin{equation}
\label{eqn:opt_move}
    \sigma_{i} =  \frac{\sum_{k=1}^{K} m_{k,i}}{||\sum_{k=1}^{K} m_{k,i}||_2}.
\end{equation}

% \begin{figure}[t!]
%     \centering
%     \includegraphics[width=2.0in]{  Figures/optimal_movement.png}
%     % \subfloat[]{{\includegraphics[width=3.4cm]{Figures/movement_rand.circles.pdf}}}%
%     % % \qquad
%     % \subfloat[]{{\includegraphics[width=3.4cm]{Figures/movement_rand.moon.pdf}}}%
%     % % \dotfill\par
%     % \quad
%     % \subfloat[]{{\includegraphics[width=3.4cm]{Figures/movement_dir.circles.pdf}}}%
%     % % \qquad
%     % \subfloat[]{{\includegraphics[width=3.4cm]{Figures/movement_dir.moon.pdf} }}%
%     \caption{Illustration of optimal movement directions. \textbf{Top row} ((a) \& (b)): Random movement directions  (green arrows), which is non-informative, uniformly sampled from two-dimensional unit sphere $\mathcal{S}^{1}$. \textbf{Bottom row} ((c) \& (d)): The optimal movement directions from SW distance between source distribution (\textcolor{green!25!black}{green}) and target distribution (\textcolor{orange}{orange}) obtained from ~Eq.~\ref{eqn:opt_move}. The movement directions are passed into the Jacobian regularization on clean samples in ~Eq.~\ref{eqn:opt_jac}.}
%     \label{fig:opt_move}
%     \vspace{-10pt}
% \end{figure}
A demonstrative example is provided in Figure~\ref{fig:opt_move_toy} to illustrate the concept of optimal trajectories. This example highlights the trajectory direction of an adversarial sample, denoted as $\tilde{x}$, within the latent space. In particular, this direction is significant, as it represents the most sensitive axis of perturbation for $x$ when exposed to adversarial interference. To address this, our approach diverges from the traditional method of random projection, as previously proposed by Hoffman et al. (2019) \cite{hoffman2019robust}. Instead, we propose to regularize the Jacobian matrix of $x$ specifically along this identified optimal trajectory. Utilizing Eq.~\ref{eqn:one_jac}, we are able to reformulate the input-output Jacobian regularization, incorporating these strategically derived projections for each sample.  The formula for this new regularization approach is presented below:  
  % For illustration, we present a toy example of optimal trajectories obtained by~Eq.~\ref{eqn:opt_move} in Fig.~\ref{fig:opt_move_toy}. The trajectory direction of an adversarial sample $\tilde{x}$ in the latent space indicates the most sensitive direction of $x$ under adversarial perturbations. Therefore, we aim to regularize the Jacobian matrix of $x$ under this optimal direction, in contrast to a random projection as was proposed in \cite{hoffman2019robust}. Using ~Eq.~\ref{eqn:one_jac}, we can rewrite the optimal input-output Jacobian regularization with previously obtained informative projections for a sample as follows:
\begin{equation}
\label{eqn:opt_jac}
    ||J(x_i | \sigma_i)||_{F}^{2} \simeq \left[ \frac{\partial (\sigma_i \cdot z_i)}{\partial x_i}\right]^{2}.
\end{equation}

Then, our overall loss function for a batch of samples $\{(x_i, y_i)\}_{\mathcal{B}}$ is expressed in the following way:
\begin{equation}
\label{eqn:overall}
 \mathcal{L} = \sum^{\mathcal{B}}_{i=1} \left( \mathcal{L_{\text{AT}}}(\tilde{x_i}, y_{i}) + \lambda_{J} ||J(x_i | \sigma_i)||_{F}^{2} \right) + \lambda_{SW}SW(\mu, \nu), 
\end{equation}
where $\mathcal{L}_{\text{AT}}$, unless stated otherwise, is cross-entropy loss of adversarial samples. In practice, sampling $K$ uniform vectors $\hat{v}_k$ can be performed simultaneously thanks to deep learning libraries. Then, the calculation of random projections and optimal trajectory steps can be vectorized and performed simultaneously.

\section{Experimental Results}
\label{sec:experiment}
%To evaluate our approach, we conducted various adversarial attacks on different datasets and two backbone networks. 
%  We first compare our proposed~\SystemName~with well-known AT baseline frameworks. Next, we show the compatibility of our \SystemName\ with different networks and high-quality datasets. Finally, we demonstrate how our \SystemName\ helps a defense model to boost its adversarial robustness via extensive ablation studies.

\subsection{Experiment Settings} 
\label{sec:exp_settings}
 We employ WideResNet34-10 \cite{zagoruyko2016wide} as our backbone architecture on two datasets~\CIFAR-10 and~\CIFAR-100 \cite{krizhevsky2009learning}. The model are trained for 100 epochs with the momentum SGD \cite{qian1999momentum} optimizer, whereas its initial learning rate is set to 0.1 and decayed by $10$ at $75^{th}$ and $90^{th}$ epoch, respectively. Adversarial samples in the training phase are generated by $L_\infty$-PGD \cite{madry2017towards} in 10 iterations with the maximal perturbation $\epsilon=8 /255$ and the perturbation step size $\eta=2/255$. For a fair comparison with different approaches \cite{pang2020bag}, we use the above settings throughout our experiments without early stopping or modifying models' architecture, and report their performances on the last epoch. For our~\SystemName~based defense models, we use the following hyper-parameter settings: $\{K=32, \lambda_J =0.002, \lambda_{SW}=64 \}$ and $\{K=128, \lambda_J =0.001, \lambda_{SW}=64 \}$ for~\CIFAR-10 and~\CIFAR-100, respectively. An ablation study of the hyper-parameters' impact on model performance is provided in Sec. \ref{abl:hyper_study} in the Appendix. In addition, we perform the basic sanity tests \cite{carlini2019evaluating} in  Sec.\ref{abl:sanity_test} to ensure that our proposed~\SystemName~does not rely on gradient obfuscation. Overall, we compare our method with four different recent SOTA AT frameworks: TRADES~\cite{zhang2019theoretically},  ALP~\cite{kannan2018adversarial}, PGD-AT~\cite{madry2017towards}, and SAT~\cite{bouniot2021optimal}. The experiments are conducted on one GeForce RTX 3090 24GB GPU with Intel Xeon Gold 6230R CPU @ 2.10GHz.

% \begin{table}[!t]
% \caption{Online fool rate (\%) \cite{mladenovic2021online} of various defense models with \CIFAR-10 and our downloaded ~\CIFAR-10-\WEB~ datasets.} 
%      \centering
%      \resizebox{0.48\textwidth}{!}{%
%      \begin{tabular}{l | c  c  c | c  c c } 
%     \Xhline{3\arrayrulewidth}
%     \multirow{2}{*}{\centering Loss}   &\multicolumn{3}{c|}{ \CIFAR-10}   &\multicolumn{3}{c}{ \CIFAR-10-\WEB}  \\
%     & {$k=100$} & {$k=200$} & {$k=500$}  & {$k=100$} & {$k=200$} & {$k=500$} \\
%     \Xhline{2\arrayrulewidth}
%     XE & $88.9_{\pm3.4}$ & $89.4_{\pm2.3}$ & $88.8_{\pm1.4}$ & $93.3_{\pm1.4}$ & $92.5_{\pm1.4}$ &$85.0_{\pm1.4}$ \Tstrut\\
%     TRADES &$74.4_{\pm4.5}$ & \textbf{73.2}$_{\pm4.3}$ & $74.0_{\pm2.0}$ &$ 78.1_{3.0}$ & $75.6 _{\pm1.4 }$ &$ 46.9_{1.4}$\Tstrut\\
%     ALP &$73.7_{\pm4.8}$ & $73.3_{\pm3.7}$ & $74.0_{\pm1.2}$ &$78.6 _{2.9}$ & $77.1 _{2.0}$ &$49.1 _{1.4}$\Tstrut\\
%     PGD-AT & $76.7_{\pm4.1}$ & $76.9_{\pm2.1}$ & $75.7_{\pm1.6}$ & $79.2 _{2.6}$ & $79.8 _{2.1}$ &$51.4 _{1.6}$\Tstrut\\
%     SAT &$75.2_{\pm3.3}$ & $74.1_{\pm2.4}$ & $73.4_{1.3}$ & $77.7 _{\pm5.4 }$ & $72.8 _{3.2}$ &$48.6 _{2.1}$\Tstrut\\
%     \rowcolor{backcolour}\SystemName &\textbf{72.9}$_{\pm2.5}$ & ${73.3}_{\pm2.3}$ & $\mathbf{73.3}_{\pm1.0}$ & \textbf{76.7}$_{3.7}$ & \textbf{72.6}$_{2.3}$ &\textbf{45.3}$_{1.6}$ \Bstrut\\
%     \Xhline{3\arrayrulewidth}
%     \end{tabular}%
%     }
%     \label{tb:online_attack}
% \end{table}

\begin{table}[!t]
\caption{Online fool rate (\%) \cite{mladenovic2021online} of various defense models with   our downloaded ~\CIFAR-10-\WEB~ and  ~\CIFAR-100-\WEB~ datasets.} 
     \centering
     \resizebox{0.46\textwidth}{!}{%
     \begin{tabular}{l | c  c  c | c  c c } 
    \Xhline{3\arrayrulewidth}
    \multirow{2}{*}{\centering Defense}   &\multicolumn{3}{c|}{ \CIFAR-10-\WEB}   &\multicolumn{3}{c}{ \CIFAR-100-\WEB}  \\
    & {$k=100$} & {$k=200$} & {$k=500$}  & {$k=100$} & {$k=200$} & {$k=500$} \\
    \Xhline{2\arrayrulewidth}
     
    TRADES   &  78.1$_{3.0}$&  75.6$_{2.1}$&  46.9$_{1.4}$ &  84.7$_{3.4}$&  85.5$_{2.2}$&  84.9$_{1.5}$
\Tstrut\\
    ALP &  78.6$_{2.9}$&  77.1$_{2.0}$&  49.1$_{1.4}$ &  85.1$_{4.6}$&  85.6$_{2.3}$&  85.3$_{1.7}$\Tstrut\\
    PGD-AT &  79.2$_{2.6}$&  79.8$_{2.1}$&  51.4$_{1.6}$ &  84.9$_{3.6}$&  86.2$_{2.9}$&  86.3$_{0.9}$\Tstrut\\
    SAT  &  77.7$_{5.4}$&  72.8$_{3.2}$&  48.6$_{2.1}$ &  86.9$_{2.6}$&  87.2 $_{1.8}$&  87.2$_{0.6}$ \Tstrut\\
    \hline
    \textit{Random JR} &  82.3$_{3.9}$&  82.5$_{2.8}$&  63.7$_{1.3}$ &  90.5$_{2.9}$&  90.1$_{1.7}$&  90.0$_{1.4}$\Tstrut\\
    \textit{SW} &  77.6$_{4.2}$&  72.0$_{2.3}$&  46.0$_{1.6}$ &  \textbf{83.9}$_{4.0}$&  85.5$_{2.6}$&  85.5$_{1.2}$\Tstrut\\
    \rowcolor{backcolour}\SystemName   &  \textbf{74.0}$_{4.3}$&  \textbf{71.3}$_{3.1}$&  \textbf{46.0}$_{1.9}$ &  84.9$_{3.8}$&  \textbf{84.6}$_{2.5}$&  \textbf{84.1}$_{1.4}$\Bstrut\\
    \Xhline{3\arrayrulewidth}
    \end{tabular}%
    }
    \label{tb:online_attack}
\end{table}
\begin{table}[!t]
\caption{Classification accuracy (\%) from defense losses integrated AWP, LBGAT, and UDR, respectively, on WRN34 with \CIFAR-100. $\ast$ indicates baseline results obtained from downloaded models. $\dagger$ indicates re-running original paper settings.} 
     \centering
     \resizebox{0.46\textwidth}{!}{%
     \begin{tabular}{l | c  c | c  c | c c } 
    \Xhline{3\arrayrulewidth}
    \multirow{2}{*}{\centering Defense}   &\multicolumn{2}{c|}{AWP$^\ast$}   &\multicolumn{2}{c|}{LBGAT$^\ast$} &\multicolumn{2}{c}{UDR$^{\dagger}$} \\
    & \textit{Clean} & AutoAtt & \textit{Clean} & AutoAtt & \textit{Clean} & AutoAtt \\
    \Xhline{2\arrayrulewidth}
    TRADES &\textit{60.17} & 28.80 & \textit{60.43}   &29.34 & 68.04 & 47.87 \Tstrut\\ % &58.17 &27.01
    \rowcolor{backcolour}\SystemName &\textit{\textbf{60.55}} & \textbf{29.79} & \textit{\textbf{62.15}}   &\textbf{29.64} &\textbf{68.31} &\textbf{49.34} \Bstrut\\ % &\textbf{58.91} &\textbf{27.12}
    \Xhline{3\arrayrulewidth}
    \end{tabular}%
    }
    \label{tb:proxy_framwork}
\end{table}
% \begin{table}[!t]
% \caption{Classification accuracy (\%) of PreActResNet18 and ResNet50 with~\ImageNet100 and~\Intel\ dataset, respectively,  under \textit{white-} and \textit{black-box} attacks.} 
% \centering
% \resizebox{0.42\textwidth}{!}{%
% \setlength{\tabcolsep}{0.5em}
% {\renewcommand{\arraystretch}{1.1}
%     \begin{tabular}{ l | c | c c | c | c } 
%     \Xhline{3\arrayrulewidth}
%     {Defense}  &{\textit{Clean}}   &MIM   &CW &Square  & AutoAtt \\
%     \Xhline{2\arrayrulewidth}
%     \multicolumn{6}{c}{~\ImageNet100 - PreActResNet18}\\
%     \hline
%     TRADES    &\textit{62.64}    &30.66   &56.95    &46.52    &24.86 \\
%     SAT$^{400}$ &\textit{45.08}    &29.08   &41.56    &36.14    &23.96 \\
%     \cellcolor{backcolour} {\SystemName}   &\cellcolor{backcolour} \textit{61.97}     &\cellcolor{backcolour} \textbf{33.20}    &\cellcolor{backcolour} \textbf{57.42}    &\cellcolor{backcolour} \textbf{49.00}    &\cellcolor{backcolour} \textbf{27.54}\\
%     \Xhline{2\arrayrulewidth}
%     \multicolumn{6}{c}{\Intel - ResNet50}\\
%     \hline
%     TRADES    &\textit{54.43}    &3.73   &{2.47}    &5.57     &0.07\\
%     SAT &\textit{50.17}    &5.93   &4.73  &6.77     &1.60\\
%     \rowcolor{backcolour} {\SystemName} &\textit{57.80}    &\textbf{8.27}  &\textbf{6.77} &\textbf{9.37}    &\textbf{3.74}\\
%     \Xhline{3\arrayrulewidth}
%     \end{tabular}%
%     }
% }
% \label{tb:appli_attack}
% \end{table}
\begin{table}[!t]
\caption{Classification accuracy (\%) of PreActResNet18 with Tiny-\ImageNet\ dataset,  under \textit{white-} and \textit{black-box} attacks.} 
\centering
\resizebox{0.46\textwidth}{!}{%
\setlength{\tabcolsep}{0.5em}
{\renewcommand{\arraystretch}{1.1}
    \begin{tabular}{ l | c  c c  c  c c |c} 
    \Xhline{3\arrayrulewidth}
    {Defense}  &{\textit{Clean}}   &MIM   &CW & FAB &Square  & AutoAtt & Avg.\\
    \Xhline{2\arrayrulewidth}
    
    TRADES   & \textit{35.91$_{\text{.10}}$}	& 11.82$_\text{.08}$	& 8.92$_\text{.13}$	&10.88$_\text{.24}$  & 15.86$_\text{.04}$	& 8.28$_\text{.09}$ & 11.15$_\text{3.00}$\\
    ALP	&\textit{39.69$_\text{.37}$}	& 8.13$_\text{.06}$	& 7.66$_\text{.08}$	&9.92$_\text{.16}$& 15.98$_\text{.34}$	& 6.48$_\text{.10}$	& 9.63$_\text{3.75}$\\
    PGD	&\textit{\underline{33.81}$_\text{.22}$}	& 11.49$_\text{.33}$	& \textbf{10.14}$_\text{.24}$	& 11.29$_\text{.32}$& 16.20$_\text{.43}$	& \textbf{8.98}$_{\text{.16}}$ & 11.60$_\text{2.77}$ \\
    SAT$^{400}$ &\multicolumn{6}{c|}{\textit{Not converge}} & -\\
    \hline
     Random JR &\textit{47.65$_\text{.19}$} &0.29$_\text{.02}$   &0.44$_\text{.04}$   &3.79$_\text{.13}$   & 5.87$_\text{.03}$   & 0.19$_\text{.01}$ & 2.12$_\text{2.59}$\\
    SW &\textit{37.05$_\text{.09}$} &12.43$_\text{.21}$  & 9.35$_\text{.19}$  & 11.19$_\text{.14}$  & 16.77$_\text{.31}$   & 8.72$_\text{.14}$ & 11.19$_\text{3.20}$\\
    \cellcolor{backcolour} {\SystemName}   &\cellcolor{backcolour} \textit{37.97$_\text{.11}$}     &\cellcolor{backcolour} \textbf{12.57}$_\text{.08}$    &\cellcolor{backcolour} {9.55}$_\text{.04}$  &\cellcolor{backcolour} \textbf{11.30}$_\text{.09}$ &\cellcolor{backcolour} \textbf{17.33}$_\text{.12}$    &\cellcolor{backcolour} {8.91}$_\text{.08}$ &\cellcolor{backcolour} \textbf{11.97}$_\text{3.34}$\\
    \Xhline{3\arrayrulewidth}
    \end{tabular}%
    }
}
\label{tb:appli_attack}
\end{table}
\subsection{Popular Adversarial Attacks}
\label{sec:adv_attacks}
 Follow \citeauthor{zhang2019theoretically}, we assess defense methods against a wide range \textit{white-box} attacks (20 iterations)\footnote{\url{https://github.com/Harry24k/adversarial-attacks-pytorch.git}}:
FGSM \cite{goodfellow2014explaining}, PGD \cite{madry2017towards}, MIM \cite{dong2018boosting}, CW$_{\infty}$ \cite{carlini2017towards}, DeepFool \cite{moosavi2016deepfool},  and FAB \cite{croce2020minimally}; and \textit{black-box} attacks (1000 iterations): Square \cite{andriushchenko2020square} and Simba \cite{guo2019simple}. We use the same parameters as \citeauthor{rice2020overfitting} for our experiments: for the $L_{\infty}$ threat model, the values of epsilon and step size are $8/255$ and  $2/255$ for CIFAR-10 and CIFAR-100, respectively. For the $L_2$ threat model, the values of epsilon and step size are 128/255 and 15/255 for all datasets. Additionally, we include AutoAttack \cite{croce2020reliable} which is a reliable adversarial evaluation framework and an ensemble of four distinct  attacks: APGD-CE, APGD-DLR \cite{croce2020reliable}, FAB \cite{croce2020minimally}, and Square \cite{andriushchenko2020square}.
The results of this experiment are presented in~Table~\ref{tb:white_attack}, where the best results are highlighted in bold. Evidently, our proposed~\SystemName~method demonstrates its superior performance across different attack paradigms. The improvement is considerable on AutoAttack, by more than $0.51\%$ and $1.21\%$ on average compared to other methods on~\CIFAR-10 and~\CIFAR-100, respectively. In addition, we include two primary components of our proposed~\SystemName, \textit{i.e.,} SW and random JR  in~Table~\ref{tb:white_attack}. While the random JR, as expected, is highly vulnerable to most of the \textit{white-box} attacks due to its adversarial-agnostic defense strategy, the SW approach is on par with other AT methods.

% \begin{figure}[t]
%     \centering
%     \subfloat[\centering ~\CIFAR-10]{{\includegraphics[width=0.46\textwidth]{  Figures/pgd_wrn28_~\CIFAR-10.pdf} }}%
%     % \qquad
%     \hfill
%     \subfloat[\centering ~\CIFAR-100]{{\includegraphics[width=0.46\textwidth]{  Figures/pgd_wrn28_~\CIFAR-100.pdf} }}%
%     \caption{  Classification accuracy (\%) of defensive WRN28 models under different number of $L_{\infty}$-PGD iterations.}%
%     \label{fig:pgd_wrn28}
% \end{figure}
 \subsection{Online Adversarial Attacks}
 \label{sec:online_attack}
In order to validate the applicability of our defense model in real-world systems, we employ the \textit{stochastic virtual} method \cite{mladenovic2021online}. This method is designed as an online attack algorithm with a transiency property; that is, an attacker makes an \textbf{irrevocable decision} regarding whether to initiate an attack or not.

For our experiment,  we curate a dataset by downloading 1,000 images and categorizing them into 10 distinct classes, mirroring the structure of the \CIFAR-10 dataset. We refer to this new subset as \CIFAR-10-\WEB. In a similar vein, we assemble a \CIFAR-100-\WEB~ dataset comprising 5,000 images on the Internet, distributed across 100 classes analogous to the \CIFAR-100 dataset. To ensure the reliability of our findings, we replicate the experiment ten times. The outcomes of these trials are detailed in Table \ref{tb:online_attack}. Notably, our novel system, \SystemName, persistently showcases the most minimal fooling rate in comparison to other methods in the \textit{k-secretary} settings \cite{mladenovic2021online}, underscoring its robustness for real-world applications.

 \subsection{Compatibility with Different Frameworks and Datasets}
 \label{sec:compatibility}

   \subsubsection{Frameworks with surrogate models and relaxed perturbations} AWP \cite{wu2020adversarial} and LBGAT \cite{cui2021learnable}  are two SOTA benchmarks boosting adversarial robustness using surrogate models during training, albeit rather computationally expensive. Furthermore, they utilized existing AT losses, such as TRADES. We integrate our proposed optimization loss into AWP and LBGAT, respectively. Additionally, we also include the results of UDR \cite{bui2022unified} that used for creating relaxed perturbed noises upon its entire distribution. We selected TRADES as the best existing loss function that was deployed with these frameworks and experimented with~\CIFAR-100-WRN34 for comparison. For consistent comparison, we use TRADES loss as our $\mathcal{L}_{\text{AT}}$ in Eq. \ref{eqn:overall}. As shown in Table~\ref{tb:proxy_framwork}, our~\SystemName\ is compatible with the three frameworks and surpasses the baseline performance by a significant margin.
  
 \subsubsection{Large scale dataset} We conducted further experiments on Tiny-\ImageNet\ with PreactResnet18, comparing our method against competitive AT methods under most challenging attacks, including MIM, CW, FAB, Square, and AutoAttack, in Table \ref{tb:appli_attack}. All defensive methods are trained in 150 epochs with same settings as in Sec. \ref{sec:exp_settings}. We note that SAT unable to converge within 400 epochs. As observed, our method not only sustains competitive performance on clean sample but also displays superior robustness against adversarial perturbations, while the second best robust model, PGD, has to sacrifice substantially its clean accuracy compared to ours (33.81\% \textit{vs.} 37.97\%). In addition, this experiment illustrates that our proposed approach does not hinder the convergence process, unlike certain alternative optimum transport methods such as SAT, when applied to datasets of significant size.

 % \subsubsection{High-quality datasets with different backbones} We extend the evaluation of \SystemName's robustness, comparing it against two leading adversarial training (AT) frameworks: TRADES and SAT, across high-quality datasets and diverse backbone architectures. Specifically, we employ the \ImageNet100 \cite{5206848} dataset with PreActResNet18 \cite{he2016identity} and the \Intel\ dataset \cite{intelanalytics} with ResNet50 \cite{he2016deep}. For the \Intel\ dataset, the experimental settings are analogous to the \CIFAR-10 configuration, albeit with an adjusted learning rate of $1e^{-3}$—decaying by $0.1$ every 7 epochs—and training spanned 50 epochs. For \ImageNet100, settings are aligned with \CIFAR-100, but with SAT's training extended to 400 epochs.  In~Table~\ref{tb:appli_attack}, we report the results with strongest attacks from~Table~\ref{tb:white_attack}, MIM, CW (\textit{white-box}) Square attack (\textit{black-box}), and AutoAttack. Remarkably, under the \SystemName\ framework, our model not only displays superior robustness against adversarial perturbations but also sustains competitive performance on clean samples. This affirms \SystemName's adaptability to various network architectures and diverse classification scenarios.

\begin{table}[!t]
\caption{ \textbf{Basic sanity tests for our ~\SystemName~ method} with \textit{white-box} PGD attack.}
\centering
\resizebox{0.96\linewidth}{!}{%
    \begin{tabular}{c|c|c|c|c|c } 
    \Xhline{3\arrayrulewidth}
    \multicolumn{6}{c }{Number of step}\\
    \hline
    \textit{Clean} &1   &10   &20  &40  &50\\
    \hline
    \textit{84.01}$_\text{.53}$  &78.86$_\text{.69}$ &56.26$_\text{.24}$ &55.38$_\text{.29}$ &55.1$_\text{.40}$ &55.08$_\text{.40}$\\
    \hline \hline
    \multicolumn{6}{c }{ Perturbation budget $\epsilon$ w/ PGD-20}\\
    \hline
    \textit{Clean} &8/255   &16/255   &24/255   &64/255  &128/255\\
    \hline
    \textit{84.01}$_\text{.53}$  &55.38$_\text{.29}$ &24.57$_\text{1.00}$ &9.66$_\text{.54}$ &0.57$_\text{.01}$ &0.00$_\text{.01}$\\
    \Xhline{3\arrayrulewidth}
    \end{tabular}%
 }
\label{tb:sanity_test}
\end{table}

\subsection{Ablation Studies and Discussions}
\subsubsection{ Sanity Tests}
\label{abl:sanity_test}
The phenomenon of gradient obfuscation arises, when a defense method is tailored such that its gradients prove ineffective for generating adversarial samples \cite{athalye2018obfuscated}. However, the method designed in that manner can be an incomplete defense to adversarial examples \cite{athalye2018obfuscated}.  Adhering to guidelines from \cite{carlini2019evaluating}, we evaluate our pre-trained model on \CIFAR10 with WRN34  to affirm  our proposed ~\SystemName~ does not lean on gradient obfuscation. As detailed in Table~\ref{tb:sanity_test}, iterative attacks
are strictly more powerful than single-step attacks, whereas when increasing perturbation budget $\epsilon$ can also raise attack successful rate. Finally,  the PGD attack attains a 100\% success rate when $\epsilon=128/255$.

\subsubsection{Loss's derivative. }
\begin{figure}[!t]
\centering
\includegraphics[width=0.48\textwidth]{ 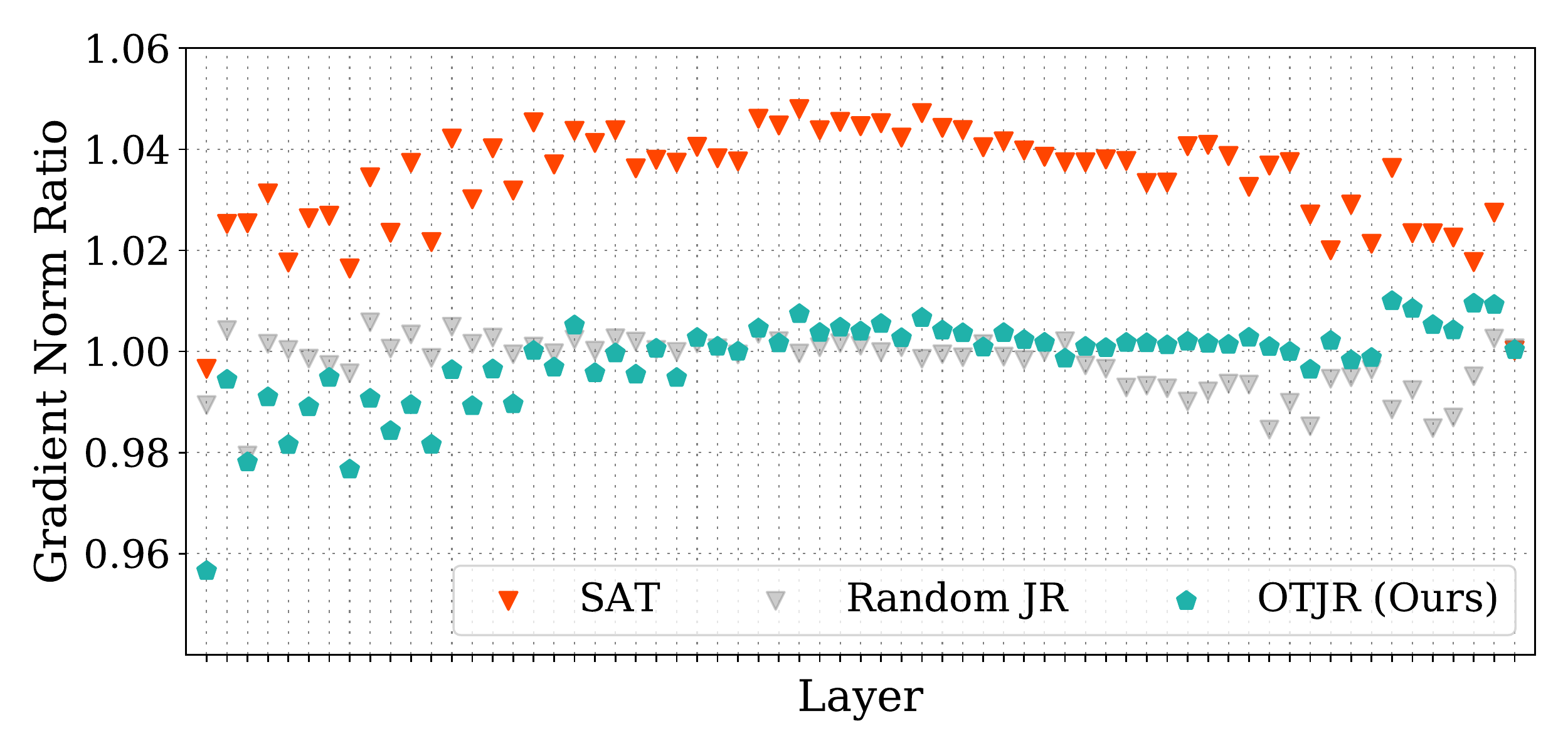}
\caption{ \textbf{Average gradient norm ratios between adversarial and clean samples.} The ratios of $\mathbb{E}(||\nabla_{\theta_i}\mathcal{L}(\tilde{x})|| / ||\nabla_{\theta_i}\mathcal{L}({x})||)$ with respect to the model's parameters $\theta_{i}$ on~\CIFAR-10 dataset.}
\label{fig:model_der_sat_JR_vs_ours}
\end{figure}

In continuation with our preliminary analysis, we highlight the disparities in defense model gradients across layers between our \SystemName\ and SAT.  Throughout intermediate layers in an attacked model, both frameworks provide stable ratios between perturbed and clean sample's gradients as shown in Fig.~\ref{fig:model_der_sat_JR_vs_ours}. It is worth noting that the gradients are derived on unobserved samples in the test set.  In the forward path, our \SystemName\ adeptly equilibrates gradients of adversarial and clean samples, with the majority of layers presenting ratio values approximating 1.   Moreover, in the backward path, since the victim model's gradients are deployed to generate more perturbations, our\ \SystemName\  model achieves better robustness when it can produce smaller gradients \textit{w.r.t.} its inputs. 
\begin{table}[!t]
\caption{\textbf{Average derivative of $\mathcal{XE}$ loss} \textit{w.r.t} input ${x}$ {at the} pixel level after various PGD iterations. The lower the derivative is, the less perturbed the adversarial samples are when the model is abused.  Best results are in \textbf{bold}.}
\centering
\resizebox{0.47\textwidth}{!}{%
    \begin{tabular}{l | r | r |r | r | r |r} 
    \Xhline{3\arrayrulewidth}
    \multirow{2}{*}{\textit{Method}}   &\multicolumn{5}{c }{$\mathbb{E}(||\nabla_{{x}} \mathcal{L})||_1)$}\\
     & \textit{Clean}  & $PGD^1$ & $PGD^5$  & $PGD^{10}$ & $PGD^{15}$ & $PGD^{20}$  \Tstrut\Bstrut\\
    \Xhline{2\arrayrulewidth}
    $\mathcal{XE}$& 854$\cdot e^{-4}$  &5492$\cdot e^{-4}$  &4894$\cdot e^{-4}$  &4852$\cdot e^{-4}$  &4975$\cdot e^{-4}$  &4900$\cdot e^{-4}$\Tstrut\\
    Random JR & 105$\cdot e^{-4}$  &149$\cdot e^{-4}$  &272$\cdot e^{-4}$  &335$\cdot e^{-4}$  &360$\cdot e^{-4}$  &370$\cdot e^{-4}$\\
    PGD & 158$\cdot e^{-4}$  &232$\cdot e^{-4}$  &420$\cdot e^{-4}$  &520$\cdot e^{-4}$  &568$\cdot e^{-4}$  &586$\cdot e^{-4}$\\
    ALP & 118$\cdot e^{-4}$  &161$\cdot e^{-4}$  &249$\cdot e^{-4}$  &298$\cdot e^{-4}$  &322$\cdot e^{-4}$  &332$\cdot e^{-4}$\\
    TRADES & 45$\cdot e^{-4}$  &53$\cdot e^{-4}$  &73$\cdot e^{-4}$  &84$\cdot e^{-4}$  &89$\cdot e^{-4}$  &90$\cdot e^{-4}$\\
   SAT & 48$\cdot e^{-4}$  &54$\cdot e^{-4}$  &65$\cdot e^{-4}$  &75$\cdot e^{-4}$  &81$\cdot e^{-4}$  &84$\cdot e^{-4}$\\
   \rowcolor{backcolour}\textbf{{\SystemName~ (\textit{Ours})}}& \textbf{43}$\cdot e^{-4}$  &\textbf{48}$\cdot e^{-4}$  &\textbf{61}$\cdot e^{-4}$  &\textbf{70}$\cdot e^{-4}$  &\textbf{73}$\cdot e^{-4}$  &\textbf{75}$\cdot e^{-4}$\Bstrut\\
    \Xhline{3\arrayrulewidth}
    \end{tabular}%
}
\label{tb:prog_input_der}
\end{table}

\begin{table*}[!t]
\caption{Comparison between ours and SAT$+$\textit{Random JR} on ~\CIFAR-10  and ~\CIFAR-100 with  WRN34.}
\centering
\resizebox{0.94\textwidth}{!}{%
    \begin{tabular}{l | l | c | c c c c c c c | c c | c } 
    \Xhline{3\arrayrulewidth}
Dataset & Defense    &\textit{Clean}    &PGD$^{20}$    &PGD$^{100}$    &$L_{2}$-PGD      &MIM &FGSM   &CW     &FAB   &Square & SimBa   &\emph{AutoAtt}\Tstrut\Bstrut\\
    \Xhline{2\arrayrulewidth}

\multirow{2}{*}{\small{~\CIFAR-10}} &   \textit{SAT+JR}  &  \textit{83.75}$_\text{.48}$	& 54.15$_\text{.17}$	& 53.87$_\text{.07}$	& 62.37$_\text{.33}$	& 54.12$_\text{.16}$	& 60.25$_\text{.22}$	& 52.22$_\text{.31}$	& 52.63$_\text{.53}$	& 61.72$_\text{.39}$	& 71.03$_\text{.65}$	& 51.15$_\text{.37}$\Tstrut\\
    &\cellcolor{backcolour} \SystemName~(\emph{\textbf{ours}}) & \cellcolor{backcolour}\textit{84.01}$_\text{.53}$ & \cellcolor{backcolour}\textbf{55.38}$_\text{.29}$ & \cellcolor{backcolour}\textbf{55.08}$_\text{.36}$ & \cellcolor{backcolour}\textbf{63.87}$_\text{.09}$ & \cellcolor{backcolour}\textbf{55.31}$_\text{.29}$ & \cellcolor{backcolour}\textbf{61.03}$_\text{.18}$ & \cellcolor{backcolour}\textbf{54.09}$_\text{.12}$ & \cellcolor{backcolour}{54.17}$_\text{.07}$ & \cellcolor{backcolour}\textbf{63.11}$_\text{.21}$ & \cellcolor{backcolour}\textbf{72.04}$_\text{.68}$ & \cellcolor{backcolour}\textbf{52.57}$_\text{.12}$\Bstrut \\
    \Xhline{3\arrayrulewidth}
\multirow{2}{*}{\small{~\CIFAR-100}} &   \textit{SAT+JR$^{400}$}  &   \textit{52.13}$_{1.55}$	& 26.18$_\text{.46}$	& 25.99$_\text{.40}$	& 32.64$_\text{.90}$	& 26.11$_\text{.43}$	& 30.14$_\text{.84}$	& 25.10$_\text{.47}$	& 25.19$_\text{.85}$	& 30.39$_{1.02}$	& 39.71$_{1.14}$	& 23.79$_\text{.42}$\Tstrut\\
     &\cellcolor{backcolour} \SystemName~(\emph{\textbf{ours}}) & \cellcolor{backcolour}\textit{58.20}$_\text{.13}$ & \cellcolor{backcolour}\textbf{32.11}$_\text{.21}$ & \cellcolor{backcolour}\textbf{32.01}$_\text{.18}$ & \cellcolor{backcolour}\textbf{43.13}$_\text{.12}$ & \cellcolor{backcolour}\textbf{32.07}$_\text{.19}$ & \cellcolor{backcolour}\textbf{34.26}$_\text{.30}$ & \cellcolor{backcolour}\textbf{29.71}$_\text{.06}$ & \cellcolor{backcolour}\textbf{29.24}$_\text{.08}$ & \cellcolor{backcolour}\textbf{36.27}$_\text{.05}$ & \cellcolor{backcolour}\textbf{49.92}$_\text{.23}$ & \cellcolor{backcolour}\textbf{28.36}$_\text{.10}$ \Bstrut \\
    \Xhline{3\arrayrulewidth}
    \end{tabular}%
}
\label{tb:naive_combine}
\end{table*}

\begin{table*}[!t]
\caption{Accuracy ($\%$) of WRN34 model trained with JR having random projections, and informative projections respectively.}
\centering
\resizebox{0.94\textwidth}{!}{%
    \begin{tabular}{l | l | c | c c c c c c c | c c | c } 
    \Xhline{3\arrayrulewidth}
Dataset & Defense    &\textit{Clean}    &PGD$^{20}$    &PGD$^{100}$    &$L_{2}$-PGD       &MIM &FGSM   &CW     &FAB   &Square & SimBa   &\emph{AutoAtt}\Tstrut\Bstrut\\
    \Xhline{2\arrayrulewidth}
\multirow{2}{*}{\small{~\CIFAR-10}} &   \textit{Random JR}     & \textit{84.99}$_\text{.14}$ & 22.67$_\text{.15}$ & 21.89$_\text{.14}$ & 60.98$_\text{.19}$ & 22.49$_\text{.18}$ & 32.99$_\text{.21}$ & 22.00$_\text{.06}$ & 21.74$_\text{.11}$ & 45.86$_\text{.20}$ & 71.20$_\text{.31}$ & 20.54$_\text{.14}$ \Tstrut\\
    &\cellcolor{backcolour} \textit{Optimal JR} &\cellcolor{backcolour} \textit{84.47}$_\text{.15}$    &\cellcolor{backcolour} \textbf{24.81}$_\text{.56}$
    &\cellcolor{backcolour} \textbf{23.97}$_\text{.48}$
   &\cellcolor{backcolour} \textbf{61.67}$_\text{.50}$
    &\cellcolor{backcolour} \textbf{24.62}$_\text{.49}$
  &\cellcolor{backcolour} \textbf{34.22}$_\text{.50}$
   &\cellcolor{backcolour} \textbf{23.91}$_\text{.70}$
   &\cellcolor{backcolour} \textbf{24.74}$_\text{.49}$
  &\cellcolor{backcolour} \textbf{47.18}$_\text{.54}$
   &\cellcolor{backcolour} \textbf{73.59}$_\text{.15}$
  &\cellcolor{backcolour} \textbf{22.84}$_\text{.64}$\Bstrut \\
    \Xhline{3\arrayrulewidth}
\multirow{2}{*}{\small{~\CIFAR-100}} &   \textit{Random JR}    & \textit{66.58}$_\text{.17}$ & 9.41$_\text{.44}$ & 8.87$_\text{.42}$ & 37.79$_\text{.31}$ & 9.27$_\text{.49}$ & 16.38$_\text{.33}$ & 10.27$_\text{.45}$ & 9.26$_\text{.23}$ & 23.53$_\text{.15}$ & 48.86$_\text{.55}$ & 8.10$_\text{.57}$ \Tstrut\\
     &\cellcolor{backcolour} \textit{Optimal JR} &\cellcolor{backcolour} \textit{65.16}$_\text{.27}$     &\cellcolor{backcolour} \textbf{10.90}$_\text{.29}$     &\cellcolor{backcolour} \textbf{10.32}$_\text{.35}$     &\cellcolor{backcolour} \textbf{38.73}$_\text{.20}$     &\cellcolor{backcolour} \textbf{10.78}$_\text{.32}$     &\cellcolor{backcolour} \textbf{17.30}$_\text{.29}$     &\cellcolor{backcolour} \textbf{11.29}$_\text{.43}$     &\cellcolor{backcolour} \textbf{10.43}$_\text{.34}$     &\cellcolor{backcolour} \textbf{24.23}$_\text{.41}$     &\cellcolor{backcolour} \textbf{50.31}$_\text{.13}$    &\cellcolor{backcolour} \textbf{9.14}$_\text{.34}$ \Bstrut \\
    \Xhline{3\arrayrulewidth}
    \end{tabular}%
}
\label{tb:randJac_vs_WDJac}
\end{table*}
In addition, in Table~\ref{tb:prog_input_der}, we present the average magnitude of cross-entropy loss's derivative \textit{w.r.t.} the input images from~\CIFAR-10 dataset with different PGD white-box attack iterations on WRN34.  Notably, as the number of attack iterations increases, the perturbation noise induced by the output loss derivatives intensifies. However, our proposed framework consistently exhibits the lowest magnitude across all iterations. This characteristic underscores the superior robustness performance of \SystemName.

% \begin{figure}[!t]
% \includegraphics[width=0.36\textwidth]{  Figures/convergence_sw_sat.pdf}
% \caption{ Convergence rates of SW and SAT on ~\CIFAR-10 and ~\CIFAR-100, respectively. While both algorithms exhibit a similar convergence on ~\CIFAR-10, SAT fails to converge on ~\CIFAR-100 in 100 epochs.} 
% \label{fig:converg}

% \end{figure}
 \subsubsection{SW distance vs. Sinkhorn divergence} Bouniot \textit{et al.} propose the SAT algorithm that deploys Sinkhorn divergence to push the distributions of clean and perturbed samples towards each other \cite{bouniot2021optimal}.
While SAT can achieve comparable results to previous research, its limitations become pronounced in high-dimensional space, i.e., datasets with a large number of classes exhibit slower convergence, as demonstrated in other research~\cite{meng2021large,petrovich2020feature}. Our empirical results indicate that \textbf{SAT struggles to converge within 100 epochs} for the \CIFAR-100 dataset. 
% as shown in Fig.~\ref{fig:converg}. 
Meanwhile, we observed that hyper-parameter settings highly affect the adversarial training~\cite{pang2020bag}, and the performance improvement by SAT can be partly achieved with additional training epochs. Nevertheless, we still include results from SAT trained in 400 epochs on ~\CIFAR-100 in \cite{bouniot2021optimal}.

 \subsubsection{Our~\SystemName~vs. Naive Combination of SAT \& random JR} To discern the impact of Jacobian regularization and distinguish our method from the naive combination of SAT and JR, we report their robustness under wide range of \textit{white-box} PGD attack in Table~\ref{tb:naive_combine}. The experiments are conducted with ~\CIFAR-10 and ~\CIFAR-100 with WRN34. Our optimal approach attains slightly better robustness on small dataset (~\CIFAR-10). On the large dataset (~\CIFAR-100), however, our  ~\SystemName~ achieves significant improvement. This phenomenon is explained by the fact that regularizing the input-output Jacobian matrix increases the difficulty of the SAT algorithm's convergence, which results in a slower convergence. Therefore, naively combining AT and random Jacobian regularization can restrain the overall optimization process.  
 
\subsubsection{Optimal vs. random projections}To verify the efficiency of the optimal Jacobian regularization, WRN34 is trained on ~\CIFAR-10 using {$\mathcal{XE}$} loss with clean samples and the regularization term using~Eq.~\ref{eqn:one_jac} and~Eq.~\ref{eqn:opt_jac} respectively, as  follow:\\
\begin{equation}
    \mathcal{L} = \sum^{\mathcal{B}}_{i=1} \left (\mathcal{L_{XE}}(x_i, y_i) + \lambda_{J}||J(x_i)||_{F}^{2} \right),
\end{equation}
where $\lambda_J$ is a hyper-parameter to balance the regularization term and {$\mathcal{XE}$} loss that is set to 0.02 for this experiment. As we can observe from Table~\ref{tb:randJac_vs_WDJac}, the Jacobian regularized model trained with SW-supported projections consistently achieves higher robustness, compared to the random projections. As shown, our proposed optimal regularization consistently achieves up to 2.5\% improvement in accuracy under \textit{AuttoAttack} compared to the random one.

\begin{figure}[!t]
     \includegraphics[width=0.52\textwidth]{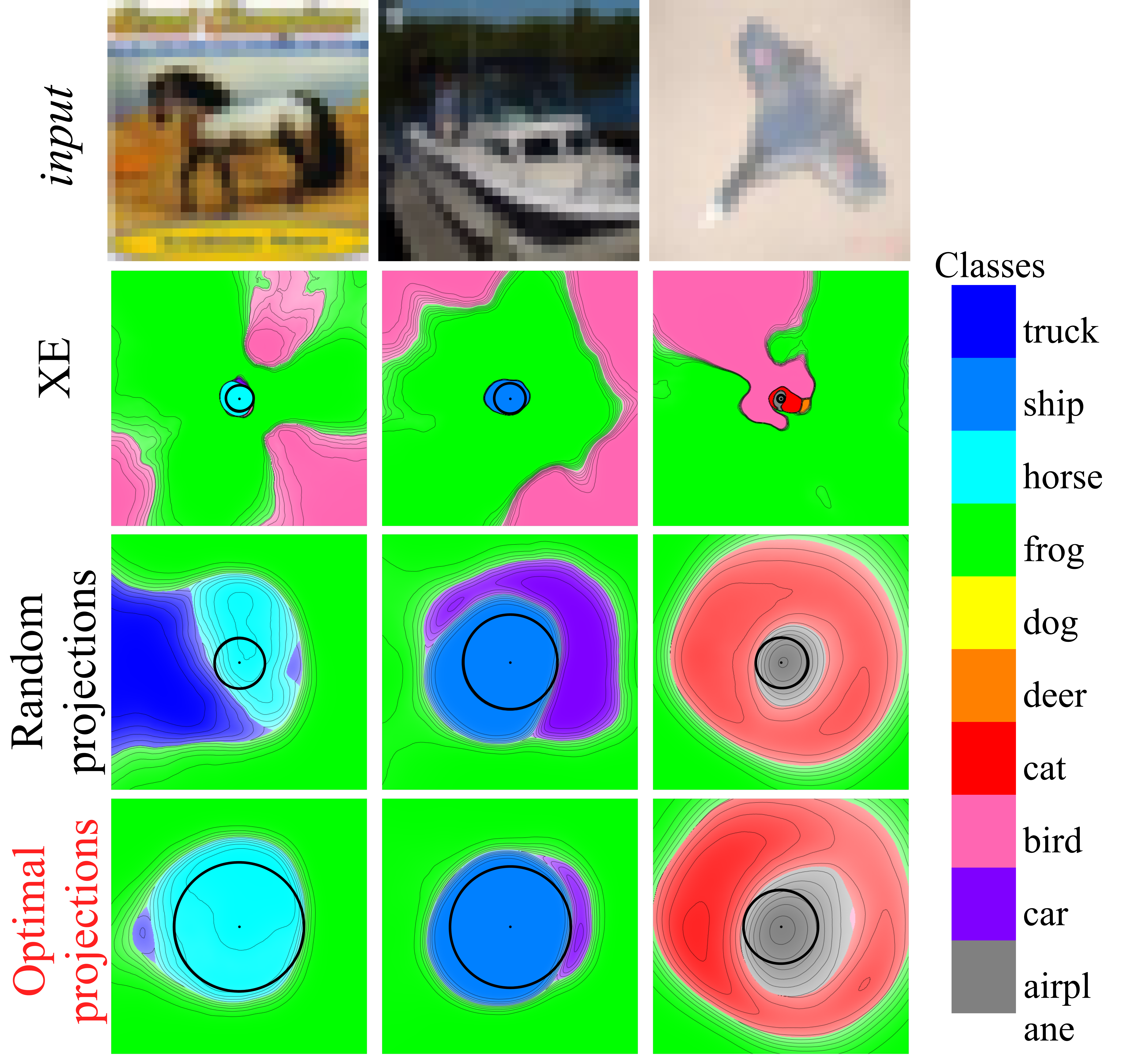}
    \caption{\textbf{Cross sections of decision boundaries in the input space}. 1st row: Sample input images. 2nd row: Model trained without regularization. 3rd row: Model train with Jacobian regularization having random projection. 4th row: Model trained with Jacobian regularization having informative projections. Our optimal JR benefits from the informative directions obtained by SW distance, and thus helps the model to regularize these sensitive direction of clean samples and produce larger decision cells.} %
    \label{fig:more_cross_section}
\end{figure}

In addition, we highlight the advantages of optimal Jacobian regularization on decision boundaries in Fig.\ref{fig:more_cross_section}. Models trained without this regularizer are notably susceptible to perturbations. However, integrating the Jacobian regularizer augments robustness by broadening the decision boundaries, evidenced by an \textbf{enlarged black circle}. Our optimal Jacobian regularizer further extends the decision boundaries, amplifying model resilience. The rationale behind this enhancement lies in the informative directions showcased in Fig.\ref{fig:opt_move_toy}, guiding the model to achieve optimal projections within the input-output Jacobian regularization framework. 

% \begin{figure*}[!t]
%   \centering
% \includegraphics[width=0.78\linewidth]{Figures/main_magnitude_activation_group.pdf}
% \caption{{Magnitude of activation} at the penultimate layer for models trained with different defense methods. Our \SystemName~ can regulate adversarial samples' magnitudes similar to clean samples' while well suppressing both of them. }%
% \label{fig:mag_activation_all}
% \end{figure*}

% \subsubsection{Activation Magnitude}
% \label{app:magnitude_activation}
% Figure ~\ref{fig:mag_activation_all} depicts the activation magnitudes at the penultimate layer of WRN34 across various AT frameworks. Although AT methods manage to bring the adversarial magnitudes closer to their clean counterparts, the magnitudes generally remain elevated, with PGD-AT being especially prominent. Through a balanced integration of the input Jacobian matrix and output distributions, our proposed method effectively mitigates the model's susceptibility to perturbed samples.

% \section{Reproducibility of Our Work}
% For reproducibility of the results presented in the paper, we share and release our source code here\footnote{https://anonymous.4open.science/r/OTJR-2024}. Upon acceptance of the paper, we plan to release the entire code. In addition, we provide more experimental results in the Appendix.

\section{Conclusion}
% We are the first to theoretically and empirically show and analyze the impacts of AT and Jacobian regularization on the robustness of DNN models, to defend against adversarial attacks. 
% In the intricate tapestry of the Web, where deep learning drives myriad services, the integrity of information is paramount. 
In the realm of deep learning, which powers a vast array of applications, maintaining the integrity of information is paramount.
In this dynamic, our research stands as the inaugural comprehensive exploration into the interplay between Adversarial Training (AT) and Jacobian regularization, especially in bolstering the robustness of Deep Neural Networks (DNNs) against adversarial forays. We show that the AT pays more attention to meaningful input samples' pixels, whereas the Jacobian regularizer agnostically silences the DNN's gradients under any perturbation from its output to input layers. Based on these characterizations, we effectively augment the AT framework by integrating input-output Jacobian matrix in order to more effectively improve the DNN's robustness. Using the optimal transport theory, our work is the first to jointly minimize the difference between the distributions of original and adversarial samples with much faster convergence. Also, the proposed SW distance produces the optimal projections for the Jacobian regularization, which can further increase the decision boundaries of a sample under perturbations, and achieves much higher performance through optimizing the best of both worlds. Our rigorous empirical evaluations, pitted against four state-of-the-art defense mechanisms across both controlled and real-world datasets, underscore the supremacy of our methodology.
% With extensive experiments comparing with four SOTA defense methods on both controlled and real-world datasets, we show that our approach outperform others. 

% \newpage

% \newpage
\bibliographystyle{ACM-Reference-Format}
\balance
\bibliography{references}


\begin{thebibliography}{47}

%%% the Web address.

\ifx \showCODEN    \undefined \def \showCODEN     #1{\unskip}     \fi
\ifx \showDOI      \undefined \def \showDOI       #1{#1}\fi
\ifx \showISBNx    \undefined \def \showISBNx     #1{\unskip}     \fi
\ifx \showISBNxiii \undefined \def \showISBNxiii  #1{\unskip}     \fi
\ifx \showISSN     \undefined \def \showISSN      #1{\unskip}     \fi
\ifx \showLCCN     \undefined \def \showLCCN      #1{\unskip}     \fi
\ifx \shownote     \undefined \def \shownote      #1{#1}          \fi
\ifx \showarticletitle \undefined \def \showarticletitle #1{#1}   \fi
\ifx \showURL      \undefined \def \showURL       {\relax}        \fi
% invisible to TeX
\providecommand\bibfield[2]{#2}
\providecommand\bibinfo[2]{#2}
\providecommand\natexlab[1]{#1}
\providecommand\showeprint[2][]{arXiv:#2}

\bibitem[Andriushchenko et~al\mbox{.}(2020)]%
        {andriushchenko2020square}
\bibfield{author}{\bibinfo{person}{Maksym Andriushchenko},
  \bibinfo{person}{Francesco Croce}, \bibinfo{person}{Nicolas Flammarion},
  {and} \bibinfo{person}{Matthias Hein}.} \bibinfo{year}{2020}\natexlab{}.
\newblock \showarticletitle{Square attack: a query-efficient black-box
  adversarial attack via random search}. In \bibinfo{booktitle}{\emph{European
  Conference on Computer Vision (ECCV)}}. Springer, \bibinfo{pages}{484--501}.
\newblock


\bibitem[Athalye et~al\mbox{.}(2018)]%
        {athalye2018obfuscated}
\bibfield{author}{\bibinfo{person}{Anish Athalye}, \bibinfo{person}{Nicholas
  Carlini}, {and} \bibinfo{person}{David Wagner}.}
  \bibinfo{year}{2018}\natexlab{}.
\newblock \showarticletitle{Obfuscated gradients give a false sense of
  security: Circumventing defenses to adversarial examples}. In
  \bibinfo{booktitle}{\emph{International conference on machine learning}}.
  PMLR, \bibinfo{pages}{274--283}.
\newblock


\bibitem[Bai et~al\mbox{.}(2021)]%
        {bai2021improving}
\bibfield{author}{\bibinfo{person}{Yang Bai}, \bibinfo{person}{Yuyuan Zeng},
  \bibinfo{person}{Yong Jiang}, \bibinfo{person}{Shu-Tao Xia},
  \bibinfo{person}{Xingjun Ma}, {and} \bibinfo{person}{Yisen Wang}.}
  \bibinfo{year}{2021}\natexlab{}.
\newblock \showarticletitle{Improving adversarial robustness via channel-wise
  activation suppressing}.
\newblock \bibinfo{journal}{\emph{International Conference on Learning
  Representations (ICLR)}} (\bibinfo{year}{2021}).
\newblock


\bibitem[Bortsova et~al\mbox{.}(2021)]%
        {bortsova2021adversarial}
\bibfield{author}{\bibinfo{person}{Gerda Bortsova}, \bibinfo{person}{Cristina
  Gonzalez-Gonzalo}, \bibinfo{person}{Suzanne Wetstein},
  \bibinfo{person}{Florian Dubost}, \bibinfo{person}{Ioannis Katramados},
  \bibinfo{person}{Laurens Hogeweg}, \bibinfo{person}{Bart Liefers},
  \bibinfo{person}{Bram Ginneken}, \bibinfo{person}{Josien Pluim},
  \bibinfo{person}{Mitko Veta}, {et~al\mbox{.}}}
  \bibinfo{year}{2021}\natexlab{}.
\newblock \showarticletitle{Adversarial attack vulnerability of medical image
  analysis systems: Unexplored factors}.
\newblock \bibinfo{journal}{\emph{Medical Image Analysis}}
  (\bibinfo{year}{2021}), \bibinfo{pages}{102141}.
\newblock


\bibitem[Bouniot et~al\mbox{.}(2021)]%
        {bouniot2021optimal}
\bibfield{author}{\bibinfo{person}{Quentin Bouniot}, \bibinfo{person}{Romaric
  Audigier}, {and} \bibinfo{person}{Angelique Loesch}.}
  \bibinfo{year}{2021}\natexlab{}.
\newblock \showarticletitle{Optimal transport as a defense against adversarial
  attacks}. In \bibinfo{booktitle}{\emph{2020 25th International Conference on
  Pattern Recognition (ICPR)}}. IEEE, \bibinfo{pages}{5044--5051}.
\newblock


\bibitem[Bui et~al\mbox{.}(2022)]%
        {bui2022unified}
\bibfield{author}{\bibinfo{person}{Tuan~Anh Bui}, \bibinfo{person}{Trung Le},
  \bibinfo{person}{Quan Tran}, \bibinfo{person}{He Zhao}, {and}
  \bibinfo{person}{Dinh Phung}.} \bibinfo{year}{2022}\natexlab{}.
\newblock \showarticletitle{A unified wasserstein distributional robustness
  framework for adversarial training}.
\newblock \bibinfo{journal}{\emph{arXiv preprint arXiv:2202.13437}}
  (\bibinfo{year}{2022}).
\newblock


\bibitem[Carlini et~al\mbox{.}(2019)]%
        {carlini2019evaluating}
\bibfield{author}{\bibinfo{person}{Nicholas Carlini}, \bibinfo{person}{Anish
  Athalye}, \bibinfo{person}{Nicolas Papernot}, \bibinfo{person}{Wieland
  Brendel}, \bibinfo{person}{Jonas Rauber}, \bibinfo{person}{Dimitris Tsipras},
  \bibinfo{person}{Ian Goodfellow}, \bibinfo{person}{Aleksander Madry}, {and}
  \bibinfo{person}{Alexey Kurakin}.} \bibinfo{year}{2019}\natexlab{}.
\newblock \showarticletitle{On evaluating adversarial robustness}.
\newblock \bibinfo{journal}{\emph{arXiv preprint arXiv:1902.06705}}
  (\bibinfo{year}{2019}).
\newblock


\bibitem[Carlini and Wagner(2017)]%
        {carlini2017towards}
\bibfield{author}{\bibinfo{person}{Nicholas Carlini} {and}
  \bibinfo{person}{David Wagner}.} \bibinfo{year}{2017}\natexlab{}.
\newblock \showarticletitle{Towards evaluating the robustness of neural
  networks}. In \bibinfo{booktitle}{\emph{2017 ieee symposium on security and
  privacy (sp)}}. IEEE, \bibinfo{pages}{39--57}.
\newblock


\bibitem[Christian et~al\mbox{.}(2013)]%
        {szegedy2013intriguing}
\bibfield{author}{\bibinfo{person}{Szegedy Christian}, \bibinfo{person}{Zaremba
  Wojciech}, \bibinfo{person}{Sutskever Ilya}, \bibinfo{person}{Bruna Joan},
  \bibinfo{person}{Erhan Dumitru}, \bibinfo{person}{Goodfellow Ian}, {and}
  \bibinfo{person}{Fergus Rob}.} \bibinfo{year}{2013}\natexlab{}.
\newblock \showarticletitle{Intriguing properties of neural networks}.
\newblock \bibinfo{journal}{\emph{arXiv preprint arXiv:1312.6199}}
  (\bibinfo{year}{2013}).
\newblock


\bibitem[Co et~al\mbox{.}(2021)]%
        {co2021jacobian}
\bibfield{author}{\bibinfo{person}{Kenneth~T Co},
  \bibinfo{person}{David~Martinez Rego}, {and} \bibinfo{person}{Emil~C Lupu}.}
  \bibinfo{year}{2021}\natexlab{}.
\newblock \showarticletitle{Jacobian Regularization for Mitigating Universal
  Adversarial Perturbations}.
\newblock \bibinfo{journal}{\emph{30th International Conference on Artificial
  Neural Networks (ICANN)}} (\bibinfo{year}{2021}).
\newblock


\bibitem[Croce and Hein(2020a)]%
        {croce2020minimally}
\bibfield{author}{\bibinfo{person}{Francesco Croce} {and}
  \bibinfo{person}{Matthias Hein}.} \bibinfo{year}{2020}\natexlab{a}.
\newblock \showarticletitle{Minimally distorted adversarial examples with a
  fast adaptive boundary attack}. In \bibinfo{booktitle}{\emph{International
  Conference on Machine Learning (ICML)}}. PMLR, \bibinfo{pages}{2196--2205}.
\newblock


\bibitem[Croce and Hein(2020b)]%
        {croce2020reliable}
\bibfield{author}{\bibinfo{person}{Francesco Croce} {and}
  \bibinfo{person}{Matthias Hein}.} \bibinfo{year}{2020}\natexlab{b}.
\newblock \showarticletitle{Reliable evaluation of adversarial robustness with
  an ensemble of diverse parameter-free attacks}. In
  \bibinfo{booktitle}{\emph{International conference on machine learning}}.
  PMLR, \bibinfo{pages}{2206--2216}.
\newblock


\bibitem[Cui et~al\mbox{.}(2021)]%
        {cui2021learnable}
\bibfield{author}{\bibinfo{person}{Jiequan Cui}, \bibinfo{person}{Shu Liu},
  \bibinfo{person}{Liwei Wang}, {and} \bibinfo{person}{Jiaya Jia}.}
  \bibinfo{year}{2021}\natexlab{}.
\newblock \showarticletitle{Learnable boundary guided adversarial training}. In
  \bibinfo{booktitle}{\emph{Proceedings of the IEEE/CVF International
  Conference on Computer Vision}}. \bibinfo{pages}{15721--15730}.
\newblock


\bibitem[Cuturi(2013)]%
        {cuturi2013sinkhorn}
\bibfield{author}{\bibinfo{person}{Marco Cuturi}.}
  \bibinfo{year}{2013}\natexlab{}.
\newblock \showarticletitle{Sinkhorn distances: Lightspeed computation of
  optimal transport}.
\newblock \bibinfo{journal}{\emph{Advances in neural information processing
  systems (NeurIPS)}}  \bibinfo{volume}{26} (\bibinfo{year}{2013}),
  \bibinfo{pages}{2292--2300}.
\newblock


\bibitem[Deng et~al\mbox{.}(2020)]%
        {deng2020analysis}
\bibfield{author}{\bibinfo{person}{Yao Deng}, \bibinfo{person}{Xi Zheng},
  \bibinfo{person}{Tianyi Zhang}, \bibinfo{person}{Chen Chen},
  \bibinfo{person}{Guannan Lou}, {and} \bibinfo{person}{Miryung Kim}.}
  \bibinfo{year}{2020}\natexlab{}.
\newblock \showarticletitle{An analysis of adversarial attacks and defenses on
  autonomous driving models}. In \bibinfo{booktitle}{\emph{2020 IEEE
  International Conference on Pervasive Computing and Communications
  (PerCom)}}. IEEE, \bibinfo{pages}{1--10}.
\newblock


\bibitem[Dong et~al\mbox{.}(2018)]%
        {dong2018boosting}
\bibfield{author}{\bibinfo{person}{Yinpeng Dong}, \bibinfo{person}{Fangzhou
  Liao}, \bibinfo{person}{Tianyu Pang}, \bibinfo{person}{Hang Su},
  \bibinfo{person}{Jun Zhu}, \bibinfo{person}{Xiaolin Hu}, {and}
  \bibinfo{person}{Jianguo Li}.} \bibinfo{year}{2018}\natexlab{}.
\newblock \showarticletitle{Boosting adversarial attacks with momentum}. In
  \bibinfo{booktitle}{\emph{Proceedings of the IEEE conference on computer
  vision and pattern recognition (CVPR)}}. \bibinfo{pages}{9185--9193}.
\newblock


\bibitem[Duchi et~al\mbox{.}(2021)]%
        {duchi2021statistics}
\bibfield{author}{\bibinfo{person}{John~C Duchi}, \bibinfo{person}{Peter~W
  Glynn}, {and} \bibinfo{person}{Hongseok Namkoong}.}
  \bibinfo{year}{2021}\natexlab{}.
\newblock \showarticletitle{Statistics of robust optimization: A generalized
  empirical likelihood approach}.
\newblock \bibinfo{journal}{\emph{Mathematics of Operations Research}}
  \bibinfo{volume}{46}, \bibinfo{number}{3} (\bibinfo{year}{2021}),
  \bibinfo{pages}{946--969}.
\newblock


\bibitem[Gao et~al\mbox{.}(2022)]%
        {gao2022wasserstein}
\bibfield{author}{\bibinfo{person}{Rui Gao}, \bibinfo{person}{Xi Chen}, {and}
  \bibinfo{person}{Anton~J Kleywegt}.} \bibinfo{year}{2022}\natexlab{}.
\newblock \showarticletitle{Wasserstein distributionally robust optimization
  and variation regularization}.
\newblock \bibinfo{journal}{\emph{Operations Research}} (\bibinfo{year}{2022}).
\newblock


\bibitem[Guo et~al\mbox{.}(2019)]%
        {guo2019simple}
\bibfield{author}{\bibinfo{person}{Chuan Guo}, \bibinfo{person}{Jacob Gardner},
  \bibinfo{person}{Yurong You}, \bibinfo{person}{Andrew~Gordon Wilson}, {and}
  \bibinfo{person}{Kilian Weinberger}.} \bibinfo{year}{2019}\natexlab{}.
\newblock \showarticletitle{Simple black-box adversarial attacks}. In
  \bibinfo{booktitle}{\emph{International Conference on Machine Learning
  (ICML)}}. PMLR, \bibinfo{pages}{2484--2493}.
\newblock


\bibitem[Helgason(2010)]%
        {helgason2010integral}
\bibfield{author}{\bibinfo{person}{Sigurdur Helgason}.}
  \bibinfo{year}{2010}\natexlab{}.
\newblock \bibinfo{booktitle}{\emph{Integral geometry and Radon transforms}}.
\newblock \bibinfo{publisher}{Springer Science \& Business Media}.
\newblock


\bibitem[Hoffman et~al\mbox{.}(2019)]%
        {hoffman2019robust}
\bibfield{author}{\bibinfo{person}{Judy Hoffman}, \bibinfo{person}{Daniel~A
  Roberts}, {and} \bibinfo{person}{Sho Yaida}.}
  \bibinfo{year}{2019}\natexlab{}.
\newblock \showarticletitle{Robust learning with Jacobian regularization}.
\newblock \bibinfo{journal}{\emph{arXiv preprint arXiv:1908.02729}}
  (\bibinfo{year}{2019}).
\newblock


\bibitem[Ian et~al\mbox{.}(2014)]%
        {goodfellow2014explaining}
\bibfield{author}{\bibinfo{person}{Goodfellow Ian}, \bibinfo{person}{Shlens
  Jonathon}, {and} \bibinfo{person}{Szegedy Christian}.}
  \bibinfo{year}{2014}\natexlab{}.
\newblock \showarticletitle{Explaining and harnessing adversarial examples}.
\newblock \bibinfo{journal}{\emph{arXiv preprint arXiv:1412.6572}}
  (\bibinfo{year}{2014}).
\newblock


\bibitem[Jakubovitz and Giryes(2018)]%
        {jakubovitz2018improving}
\bibfield{author}{\bibinfo{person}{Daniel Jakubovitz} {and}
  \bibinfo{person}{Raja Giryes}.} \bibinfo{year}{2018}\natexlab{}.
\newblock \showarticletitle{Improving DNN robustness to adversarial attacks
  using Jacobian regularization}. In \bibinfo{booktitle}{\emph{Proceedings of
  the European Conference on Computer Vision (ECCV)}}.
  \bibinfo{pages}{514--529}.
\newblock


\bibitem[Kannan et~al\mbox{.}(2018)]%
        {kannan2018adversarial}
\bibfield{author}{\bibinfo{person}{Harini Kannan}, \bibinfo{person}{Alexey
  Kurakin}, {and} \bibinfo{person}{Ian Goodfellow}.}
  \bibinfo{year}{2018}\natexlab{}.
\newblock \showarticletitle{Adversarial logit pairing}.
\newblock \bibinfo{journal}{\emph{arXiv preprint arXiv:1803.06373}}
  (\bibinfo{year}{2018}).
\newblock


\bibitem[Krizhevsky et~al\mbox{.}(2009)]%
        {krizhevsky2009learning}
\bibfield{author}{\bibinfo{person}{Alex Krizhevsky}, \bibinfo{person}{Geoffrey
  Hinton}, {et~al\mbox{.}}} \bibinfo{year}{2009}\natexlab{}.
\newblock \bibinfo{booktitle}{\emph{Learning multiple layers of features from
  tiny images}}.
\newblock \bibinfo{type}{{T}echnical {R}eport}.
  \bibinfo{institution}{University of Toronto, Toronto}.
\newblock


\bibitem[Kuhn et~al\mbox{.}(2019)]%
        {kuhn2019wasserstein}
\bibfield{author}{\bibinfo{person}{Daniel Kuhn},
  \bibinfo{person}{Peyman~Mohajerin Esfahani}, \bibinfo{person}{Viet~Anh
  Nguyen}, {and} \bibinfo{person}{Soroosh Shafieezadeh-Abadeh}.}
  \bibinfo{year}{2019}\natexlab{}.
\newblock \showarticletitle{Wasserstein distributionally robust optimization:
  Theory and applications in machine learning}.
\newblock In \bibinfo{booktitle}{\emph{Operations research \& management
  science in the age of analytics}}. \bibinfo{publisher}{Informs},
  \bibinfo{pages}{130--166}.
\newblock


\bibitem[Kurakin et~al\mbox{.}(2016)]%
        {kurakin2016adversarial}
\bibfield{author}{\bibinfo{person}{Alexey Kurakin}, \bibinfo{person}{Ian
  Goodfellow}, \bibinfo{person}{Samy Bengio}, {et~al\mbox{.}}}
  \bibinfo{year}{2016}\natexlab{}.
\newblock \bibinfo{title}{Adversarial examples in the physical world}.
\newblock
\newblock


\bibitem[Ma et~al\mbox{.}(2021)]%
        {ma2021understanding}
\bibfield{author}{\bibinfo{person}{Xingjun Ma}, \bibinfo{person}{Yuhao Niu},
  \bibinfo{person}{Lin Gu}, \bibinfo{person}{Yisen Wang},
  \bibinfo{person}{Yitian Zhao}, \bibinfo{person}{James Bailey}, {and}
  \bibinfo{person}{Feng Lu}.} \bibinfo{year}{2021}\natexlab{}.
\newblock \showarticletitle{Understanding adversarial attacks on deep learning
  based medical image analysis systems}.
\newblock \bibinfo{journal}{\emph{Pattern Recognition}}  \bibinfo{volume}{110}
  (\bibinfo{year}{2021}), \bibinfo{pages}{107332}.
\newblock


\bibitem[Madry et~al\mbox{.}(2018)]%
        {madry2017towards}
\bibfield{author}{\bibinfo{person}{Aleksander Madry},
  \bibinfo{person}{Aleksandar Makelov}, \bibinfo{person}{Ludwig Schmidt},
  \bibinfo{person}{Dimitris Tsipras}, {and} \bibinfo{person}{Adrian Vladu}.}
  \bibinfo{year}{2018}\natexlab{}.
\newblock \showarticletitle{Towards deep learning models resistant to
  adversarial attacks}.
\newblock \bibinfo{journal}{\emph{International Conference on Learning
  Representations (ICLR)}} (\bibinfo{year}{2018}).
\newblock


\bibitem[Meng et~al\mbox{.}(2019)]%
        {meng2021large}
\bibfield{author}{\bibinfo{person}{Cheng Meng}, \bibinfo{person}{Yuan Ke},
  \bibinfo{person}{Jingyi Zhang}, \bibinfo{person}{Mengrui Zhang},
  \bibinfo{person}{Wenxuan Zhong}, {and} \bibinfo{person}{Ping Ma}.}
  \bibinfo{year}{2019}\natexlab{}.
\newblock \showarticletitle{Large-scale optimal transport map estimation using
  projection pursuit}.
\newblock \bibinfo{journal}{\emph{Advances in Neural Information Processing
  Systems (NeurIPS)}} (\bibinfo{year}{2019}).
\newblock


\bibitem[Mladenovic et~al\mbox{.}(2021)]%
        {mladenovic2021online}
\bibfield{author}{\bibinfo{person}{Andjela Mladenovic}, \bibinfo{person}{Joey
  Bose}, \bibinfo{person}{William~L Hamilton}, \bibinfo{person}{Simon
  Lacoste-Julien}, \bibinfo{person}{Pascal Vincent}, \bibinfo{person}{Gauthier
  Gidel}, {et~al\mbox{.}}} \bibinfo{year}{2021}\natexlab{}.
\newblock \showarticletitle{Online Adversarial Attacks}. In
  \bibinfo{booktitle}{\emph{International Conference on Learning
  Representations}}.
\newblock


\bibitem[Moosavi-Dezfooli et~al\mbox{.}(2016)]%
        {moosavi2016deepfool}
\bibfield{author}{\bibinfo{person}{Seyed-Mohsen Moosavi-Dezfooli},
  \bibinfo{person}{Alhussein Fawzi}, {and} \bibinfo{person}{Pascal Frossard}.}
  \bibinfo{year}{2016}\natexlab{}.
\newblock \showarticletitle{Deepfool: a simple and accurate method to fool deep
  neural networks}. In \bibinfo{booktitle}{\emph{Proceedings of the IEEE
  conference on computer vision and pattern recognition}}.
  \bibinfo{pages}{2574--2582}.
\newblock


\bibitem[Pang et~al\mbox{.}(2020)]%
        {pang2020bag}
\bibfield{author}{\bibinfo{person}{Tianyu Pang}, \bibinfo{person}{Xiao Yang},
  \bibinfo{person}{Yinpeng Dong}, \bibinfo{person}{Hang Su}, {and}
  \bibinfo{person}{Jun Zhu}.} \bibinfo{year}{2020}\natexlab{}.
\newblock \showarticletitle{Bag of tricks for adversarial training}.
\newblock \bibinfo{journal}{\emph{arXiv preprint arXiv:2010.00467}}
  (\bibinfo{year}{2020}).
\newblock


\bibitem[Petrovich et~al\mbox{.}(2020)]%
        {petrovich2020feature}
\bibfield{author}{\bibinfo{person}{Mathis Petrovich}, \bibinfo{person}{Chao
  Liang}, \bibinfo{person}{Ryoma Sato}, \bibinfo{person}{Yanbin Liu},
  \bibinfo{person}{Yao-Hung~Hubert Tsai}, \bibinfo{person}{Linchao Zhu},
  \bibinfo{person}{Yi Yang}, \bibinfo{person}{Ruslan Salakhutdinov}, {and}
  \bibinfo{person}{Makoto Yamada}.} \bibinfo{year}{2020}\natexlab{}.
\newblock \showarticletitle{Feature robust optimal transport for
  high-dimensional data}.
\newblock \bibinfo{journal}{\emph{arXiv preprint arXiv:2005.12123}}
  (\bibinfo{year}{2020}).
\newblock


\bibitem[Qian(1999)]%
        {qian1999momentum}
\bibfield{author}{\bibinfo{person}{Ning Qian}.}
  \bibinfo{year}{1999}\natexlab{}.
\newblock \showarticletitle{On the momentum term in gradient descent learning
  algorithms}.
\newblock \bibinfo{journal}{\emph{Neural networks}} \bibinfo{volume}{12},
  \bibinfo{number}{1} (\bibinfo{year}{1999}), \bibinfo{pages}{145--151}.
\newblock


\bibitem[Rahimian and Mehrotra(2019)]%
        {rahimian2019distributionally}
\bibfield{author}{\bibinfo{person}{Hamed Rahimian} {and}
  \bibinfo{person}{Sanjay Mehrotra}.} \bibinfo{year}{2019}\natexlab{}.
\newblock \showarticletitle{Distributionally robust optimization: A review}.
\newblock \bibinfo{journal}{\emph{arXiv preprint arXiv:1908.05659}}
  (\bibinfo{year}{2019}).
\newblock


\bibitem[Rice et~al\mbox{.}(2020)]%
        {rice2020overfitting}
\bibfield{author}{\bibinfo{person}{Leslie Rice}, \bibinfo{person}{Eric Wong},
  {and} \bibinfo{person}{Zico Kolter}.} \bibinfo{year}{2020}\natexlab{}.
\newblock \showarticletitle{Overfitting in adversarially robust deep learning}.
  In \bibinfo{booktitle}{\emph{International Conference on Machine Learning}}.
  PMLR, \bibinfo{pages}{8093--8104}.
\newblock


\bibitem[Rony et~al\mbox{.}(2019)]%
        {rony2019decoupling}
\bibfield{author}{\bibinfo{person}{J{\'e}r{\^o}me Rony},
  \bibinfo{person}{Luiz~G Hafemann}, \bibinfo{person}{Luiz~S Oliveira},
  \bibinfo{person}{Ismail~Ben Ayed}, \bibinfo{person}{Robert Sabourin}, {and}
  \bibinfo{person}{Eric Granger}.} \bibinfo{year}{2019}\natexlab{}.
\newblock \showarticletitle{Decoupling direction and norm for efficient
  gradient-based l2 adversarial attacks and defenses}. In
  \bibinfo{booktitle}{\emph{Proceedings of the IEEE/CVF Conference on Computer
  Vision and Pattern Recognition (CVPR)}}. \bibinfo{pages}{4322--4330}.
\newblock


\bibitem[Shafahi et~al\mbox{.}(2019)]%
        {shafahi2019adversarial}
\bibfield{author}{\bibinfo{person}{Ali Shafahi}, \bibinfo{person}{Mahyar
  Najibi}, \bibinfo{person}{Amin Ghiasi}, \bibinfo{person}{Zheng Xu},
  \bibinfo{person}{John Dickerson}, \bibinfo{person}{Christoph Studer},
  \bibinfo{person}{Larry~S Davis}, \bibinfo{person}{Gavin Taylor}, {and}
  \bibinfo{person}{Tom Goldstein}.} \bibinfo{year}{2019}\natexlab{}.
\newblock \showarticletitle{Adversarial training for free!}
\newblock \bibinfo{journal}{\emph{arXiv preprint arXiv:1904.12843}}
  (\bibinfo{year}{2019}).
\newblock


\bibitem[Shafieezadeh~Abadeh et~al\mbox{.}(2015)]%
        {shafieezadeh2015distributionally}
\bibfield{author}{\bibinfo{person}{Soroosh Shafieezadeh~Abadeh},
  \bibinfo{person}{Peyman~M Mohajerin~Esfahani}, {and} \bibinfo{person}{Daniel
  Kuhn}.} \bibinfo{year}{2015}\natexlab{}.
\newblock \showarticletitle{Distributionally robust logistic regression}.
\newblock \bibinfo{journal}{\emph{Advances in Neural Information Processing
  Systems}}  \bibinfo{volume}{28} (\bibinfo{year}{2015}).
\newblock


\bibitem[Vialard(2019)]%
        {vialard2019elementary}
\bibfield{author}{\bibinfo{person}{Fran{\c{c}}ois-Xavier Vialard}.}
  \bibinfo{year}{2019}\natexlab{}.
\newblock \showarticletitle{An elementary introduction to entropic
  regularization and proximal methods for numerical optimal transport}.
\newblock  (\bibinfo{year}{2019}).
\newblock


\bibitem[Villani(2008)]%
        {villani2008optimal}
\bibfield{author}{\bibinfo{person}{C{\'e}dric Villani}.}
  \bibinfo{year}{2008}\natexlab{}.
\newblock \bibinfo{booktitle}{\emph{Optimal transport: old and new}}.
  Vol.~\bibinfo{volume}{338}.
\newblock \bibinfo{publisher}{Springer Science \& Business Media}.
\newblock


\bibitem[Wu et~al\mbox{.}(2020)]%
        {wu2020adversarial}
\bibfield{author}{\bibinfo{person}{Dongxian Wu}, \bibinfo{person}{Shu-Tao Xia},
  {and} \bibinfo{person}{Yisen Wang}.} \bibinfo{year}{2020}\natexlab{}.
\newblock \showarticletitle{Adversarial weight perturbation helps robust
  generalization}.
\newblock \bibinfo{journal}{\emph{Advances in Neural Information Processing
  Systems}}  \bibinfo{volume}{33} (\bibinfo{year}{2020}),
  \bibinfo{pages}{2958--2969}.
\newblock


\bibitem[Zagoruyko and Komodakis(2016)]%
        {zagoruyko2016wide}
\bibfield{author}{\bibinfo{person}{Sergey Zagoruyko} {and}
  \bibinfo{person}{Nikos Komodakis}.} \bibinfo{year}{2016}\natexlab{}.
\newblock \showarticletitle{Wide residual networks}.
\newblock \bibinfo{journal}{\emph{arXiv preprint arXiv:1605.07146}}
  (\bibinfo{year}{2016}).
\newblock


\bibitem[Zhang and Wang(2019)]%
        {zhang2019defense}
\bibfield{author}{\bibinfo{person}{Haichao Zhang} {and} \bibinfo{person}{Jianyu
  Wang}.} \bibinfo{year}{2019}\natexlab{}.
\newblock \showarticletitle{Defense against adversarial attacks using feature
  scattering-based adversarial training}.
\newblock \bibinfo{journal}{\emph{Advances in Neural Information Processing
  Systems (NeurIPS)}}  \bibinfo{volume}{32} (\bibinfo{year}{2019}),
  \bibinfo{pages}{1831--1841}.
\newblock


\bibitem[Zhang et~al\mbox{.}(2019)]%
        {zhang2019theoretically}
\bibfield{author}{\bibinfo{person}{Hongyang Zhang}, \bibinfo{person}{Yaodong
  Yu}, \bibinfo{person}{Jiantao Jiao}, \bibinfo{person}{Eric Xing},
  \bibinfo{person}{Laurent El~Ghaoui}, {and} \bibinfo{person}{Michael Jordan}.}
  \bibinfo{year}{2019}\natexlab{}.
\newblock \showarticletitle{Theoretically principled trade-off between
  robustness and accuracy}. In \bibinfo{booktitle}{\emph{International
  Conference on Machine Learning (ICML)}}. PMLR, \bibinfo{pages}{7472--7482}.
\newblock


\bibitem[Zhang et~al\mbox{.}(2020)]%
        {zhang2020geometry}
\bibfield{author}{\bibinfo{person}{Jingfeng Zhang}, \bibinfo{person}{Jianing
  Zhu}, \bibinfo{person}{Gang Niu}, \bibinfo{person}{Bo Han},
  \bibinfo{person}{Masashi Sugiyama}, {and} \bibinfo{person}{Mohan
  Kankanhalli}.} \bibinfo{year}{2020}\natexlab{}.
\newblock \showarticletitle{Geometry-aware instance-reweighted adversarial
  training}.
\newblock \bibinfo{journal}{\emph{International Conference on Learning
  Representations (ICLR)}} (\bibinfo{year}{2020}).
\newblock


\end{thebibliography}
\newpage

% \setcounter{page}{1}
% \newcount\cvprrulercount
% \setcounter{cvprrulercount}{1}
% \begin{center}
% \vspace{30pt}Supplementary Material for\\\textbf{OTJR: Optimal Transport Meets Optimal Jacobian Regularization for Adversarial Robustness\\}\vspace{16pt} 
% \end{center}

{\huge\textbf{Appendix}}
\appendix
\renewcommand{\thesection}{\Alph{section}}

\section{Threat Model and Adversarial Sample}
In this section, we summarize essential terminologies of adversarial settings related to our work.  We first define a threat model, which consists of a set of assumptions about the adversary. Then, we describe the generation mechanism of adversarial samples in AT frameworks for the threat model defending against adversarial attacks.

\subsection{Threat Model}
Adversarial perturbation was firstly discovered by~\citeauthor{szegedy2013intriguing}, and it instantly strikes an array of studies in both adversarial attack and adversarial robustness. ~\citeauthor{carlini2019evaluating} specifies a threat model for evaluating a defense method including a set of assumptions about the adversary's goals, capabilities, and knowledge, which are briefly delineated as follows:
 \begin{itemize}
     \item Adversary's goals could be either simply deceiving a model to make the wrong prediction to any classes from a perturbed input or making the model misclassify a specific class to an intended class. They are known as \textit{untargeted} and \textit{targeted} modes, respectively.
     \item Adversary's capabilities define reasonable constraints imposed on the attackers. For instance, a $L_p$ certified robust model is determined with the worst-case loss function $\mathcal{L}$ for a given perturbation budget $\epsilon$:
     \begin{equation}
         \mathbb{E}_{(x,y)\sim \mathcal{D}} \Bigg[\max_{ \tilde{x}\in {B}_{p} (x, \epsilon)} \mathcal{L}(f(\tilde{x}), y) \Bigg],
     \end{equation}
     where ${B}_{p} (x, \epsilon)=\{u \in \mathbb{R}^{\mathcal{I}}: ||u-x||_{p} \leq  \epsilon\}$.
     \item Adversary's knowledge indicates what knowledge of the threat model that an attacker is assumed to have. Typically, \textit{white-box} and \textit{black-box} attacks are two most popular scenarios studied. The white-box settings assume that attackers have full knowledge of the model's parameters and its defensive scheme. In contrast, the black-box settings have varying degrees of access to the model's parameter or the defense.
 \end{itemize}
 
Bearing these assumptions about the adversary, we describe how a defense model generates adversarial samples for its training in the following section.   

 \subsection{Adversarial Sample in AT}
 Among multiple attempts to defend against adversarial perturbed samples, adversarial training (AT) is known as the most successful defense method. In fact, AT is an data-augmenting training method that originates  from the work of \cite{goodfellow2014explaining}, where crafted adversarial samples are created by the fast gradient sign method (FGSM), and mixed into the mini-batch training data. Subsequently, a wide range of studies focus on developing powerful attacks \cite{kurakin2016adversarial,dong2018boosting,carlini2017towards,croce2020minimally}. Meanwhile, in the opposite direction to the adversarial attack, there are also several attempts to resist against adversarial examples \cite{kannan2018adversarial,zhang2019defense,shafahi2019adversarial}. In general, a defense model is optimized by solving a minimax problem:
  \begin{equation}
      \min _{\theta} \Big[\max_{\tilde{x} \in {B}_{p}(x, \epsilon)} \mathcal{L_{XE}}(\tilde{x}, y; \theta) \Big ],
  \end{equation}
where the inner maximization tries to successfully create perturbation samples subjected to an $\epsilon$-radius ball $B$ around the clean sample $x$ in $L_p$ space. The outer minimization tries to adjust the model's parameters to minimize the loss caused by the inner attacks. Among existing defensive AT, PGD-AT \cite{madry2017towards} becomes the most popular one, in which the inner maximization is approximated by the multi-step projected gradient (PGD) method:
 \begin{equation}
 \label{eqn:pgd_gen}
     \tilde{x}_{t+1} = \Pi_{x}^{\epsilon}(\tilde{x_{t}} +\eta \cdot \text{sgn} (\nabla _{\tilde{x_{t}}} \mathcal{L}(\tilde{x_{t}}, y))),
 \end{equation}
 where $\Pi_{x}^{\epsilon}$ is an operator that projects its input into the feasible region ${B}_{\infty}(x, \epsilon)$, and $\eta \in \mathbb{R}$ is called step size. The loss function in Eq.~\ref{eqn:pgd_gen} can be modulated to derive different variants of generation mechanism for adversarial samples in AT. For example, Zhang \textit{et al.} \cite{zhang2019theoretically} utilizes the loss between the likelihood of clean and adversarial samples for updating the adversarial samples. In our work, we use Eq.~\ref{eqn:pgd_gen} as our generation mechanism for our AT framework.

\begin{algorithm*}[t!]
\caption{\SystemName: AT with SW and optimal Jacobian regularization}
\label{alg:at_infor_reg}
\begin{algorithmic}[1]
\Require DNN $f$
parameterized by $\theta$, training dataset $\mathcal{D}$. Number of projection $K$. Maximum perturbation $\epsilon$, step size $\eta$, number of adversarial iteration $P$. Loss' hyper-parameters $\lambda_J$ and $\lambda_{SW}$. Learning rate $\alpha$ and a mini-batch size of $\mathcal{B}$.
\While{not converged}
    \For{$\{(x_{i},\textrm{ } y_{i})\}_{\mathcal{B}} \in \mathcal{D}$}
        \State $\nu :  = z_{i} = f_{\theta}(x_i)| _{i=1,..,\mathcal{B}}$ \Comment{\textit{forward a batch of clean samples through the model}}
        \For{iteration $t \gets 1$ to $P$}
            \State $\tilde{x_i} = \Pi_{x}^{\epsilon}(\tilde{x_i} +\eta \cdot \text{sgn} (\nabla _{\tilde{x_i}} \mathcal{L}(\tilde{x_i}, y_i)))| _{i=1,..,\mathcal{B}}$  \Comment{{\emph{generate adv. samples by $L_{\infty}$-PGD in $P$ iterations}}}
        \EndFor
        \State $\mu :  = \tilde{z_i} = f_{\theta}(\tilde{x_i})| _{i=1,..,\mathcal{B}}$  \Comment{\textit{forward a batch of adv. samples through the model}}
        
        \State $SW \gets 0$ \Comment{\emph{initialize SW loss}}
        \State $\sigma_i \gets 0|_{i=1,..,\mathcal{B}}$ \Comment{\emph{initialize $\mathcal{B}$ Jacobian projections}}
        \For{iteration $k \gets 1$ to $K$}
            \State $\hat{v}_k$ $\gets$  $\mathcal{U}(\mathcal{S}^{C-1})$ \Comment{{\emph{uniformly sample $\hat{v}_k$ from $\mathcal{S}^{C-1}$}}}
            
            \State $SW \gets SW + \psi  \left(\tau_1 \circ \mathcal{R}_{\hat{v}_k}\mu, \tau_2 \circ\mathcal{R}_{\hat{v}_k}\nu \right)$ \Comment{{\emph{add SW under projection $\hat{v}_k$}}}
            \State $m_k \gets \left(\tau_1^{-1} \circ \tau_2 \circ \mathcal{R}_{\hat{v}_k}\nu - \mathcal{R}_{\hat{v}_k}\mu \right) \otimes \hat{v}_k$ \Comment{{\emph{calculate samples' movements under $\hat{v}_k$}}}
            \State $\sigma_i \gets \sigma_i + m_{k,i}|_{i=1,..,\mathcal{B}}$
        \EndFor
        \State $\sigma_i \gets \sigma_i / ||\sigma_i||_2 $  $ |_{i=1,..,\mathcal{B}}$
        \State $\mathcal{L} \gets \sum^{\mathcal{B}} \left( \mathcal{L_{XE}}(\tilde{x_i}, y_i) + \lambda_J ||J(x_i | \sigma_i)||_{F}^{2} \right)  $  $ +\quad \lambda_{SW}SW(\mu, \nu) $ \Comment{{\emph{overall loss}}}
        \State $\theta \gets \theta - \alpha \cdot \nabla _{\theta} \mathcal{L}$ \Comment{{\emph{update model's parameters $\theta$}}}
    \EndFor
\EndWhile
\end{algorithmic}
\label{alg:opt_move}
\end{algorithm*}

% \section{Appendix}
% In this section, we first briefly present our algorithm for training our \SystemName framework. Next, we conduct intensive experiments on various AT methods. 
\section{Training Algorithm for \SystemName}

Our end-to-end algorithm for optimizing ~Eq.~\ref{eqn:overall} is provided in  Algorithm \ref{alg:opt_move}. As mentioned, in practice, deep learning libraries allow for the simultaneous sampling of $K$ uniform vectors, denoted as $\hat{v}_k$. Consequently, the computation of random projections and the determination of optimal movement steps can be effectively vectorized and executed concurrently.

\section{Hyper-parameter sensitivity}
\label{abl:hyper_study}
In Table~\ref{tb:ab_lambda_sw}, we present ablation studies focusing on hyper-parameter sensitivities, namely, $\lambda_J$, $\lambda_{SW}$, and $K$, using the \CIFAR-10 dataset and WRN34 architecture. We observe that excessive $\lambda_J$ values compromise accuracy and robustness, a result of the loss function gradients inducing adversarial perturbations during the AT step. While $\lambda_{SW}$ offers flexibility in selection, models with minimal $\lambda_{SW}$ values inadequately address adversarial samples, and high values risk eroding clean accuracy. For the slice count $K$, a lower count fails to encapsulate transportation costs across latent space distributions; conversely, an overly large $K$ brings marginal benefits at the expense of extended training times. We acknowledge potential gains from further hyper-parameter optimization.
\section{Training Time}
Table~\ref{tb:training_time} indicates the average training time per epoch of all AT methods on our machine architecture using WRN34 model on ~\CIFAR-100 dataset. Notably, although the SAT algorithm demonstrates a commendable per-epoch training duration, its convergence necessitates up to four times more epochs than alternative methods, especially on large scale datasets such as CIFAR-100. Despite our method delivering notable enhancements over prior state-of-the-art frameworks, its computational demand during training remains within acceptable bounds.
\begin{table}[th!]
\caption{\textbf{Hyper-parameter tuning.} The sensitivities of hyper-parameters: $\lambda_J$, $\lambda_{SW}$ and $K$. Without Jacobian regularization, the model cannot achieve the best performance. Trade-off between model's accuracy vs. robustness is shown via $\lambda_{SW}$.}
\centering
\resizebox{0.44\textwidth}{!}{%
    \begin{tabular}{c | c | c | c | c | c | c } 
    \Xhline{3\arrayrulewidth}
    \multicolumn{3}{c | }{Hyper-parameters}    &\multicolumn{4}{c}{Robustness}\\
    \Xhline{1\arrayrulewidth}
    $\lambda_J$    &$\lambda_{SW}$ &$K$   &\textit{Clean}   &PGD$^{20}$    &PGD$^{100}$    &AutoAttack \\
    \hline \hline
     \rowcolor{backcolour}  0.002    &64    &32     &\textit{84.53}    &55.07    &54.69    &52.41\\
     \hline
    \textcolor{blue}{\underline{0.01}}    &64    &32  &\textit{84.75}    &54.37    &54.06    &52.13 \\
    \textcolor{blue}{\underline{0.05}}    &64    &32  &\textit{82.82}    &54.98    &54.72    &52.00\\
    \Xhline{2\arrayrulewidth}
    0.002    &\textcolor{blue}{\underline{32}} &32     &\textit{85.47}    &54.85    &54.46    &52.23\\
    0.002    &\textcolor{blue}{\underline{72}}    &32   &\textit{83.19}    &55.70    &55.40    &53.04\\
    \Xhline{2\arrayrulewidth}
    0.002    &64    &\textcolor{blue}{\underline{16}}  &\textit{81.47}    &55.10    &54.98    &51.82\\
    0.002    &64   &\textcolor{blue}{\underline{64}}  &\textit{85.79}    &53.80    &53.36    &51.83\\
    \Xhline{3\arrayrulewidth}
    \end{tabular}%
}
\label{tb:ab_lambda_sw}
\end{table}

\begin{table}[th!]
\caption{\textbf{Training time per epoch} of AT methods. Even though our method's training time/epoch is slightly slower than the SAT's as the additional Jacobian regularization, it can achieve faster convergence on large-scale datasets. }
\centering
\resizebox{0.86\linewidth}{!}{%
    \begin{tabular}{l | c || l | c } 
    \Xhline{2\arrayrulewidth}
    Method & Time (\textit{mins}) &Method & Time (\textit{mins})
     \\
    \hline \hline
    $\mathcal{XE}$ &1.63$_\text{.00}$ &PGD-AT &12.32$_\text{.03}$ \\
    ALP & 13.56$_\text{.03}$  & TRADES &16.42$_\text{.06}$ \\
      SAT &14.68$_\text{.01}$ &{\SystemName~(Ours)} &18.02$_\text{.12}$ \\
    \Xhline{2\arrayrulewidth}
    \end{tabular}%
}
\label{tb:training_time}
\end{table}

\section{Activation Magnitude}
\label{app:magnitude_activation}
\begin{figure*}[t]
    \centering
    \subfloat[\centering $\mathcal{XE}$ ]{{\includegraphics[width=0.42\linewidth]{ 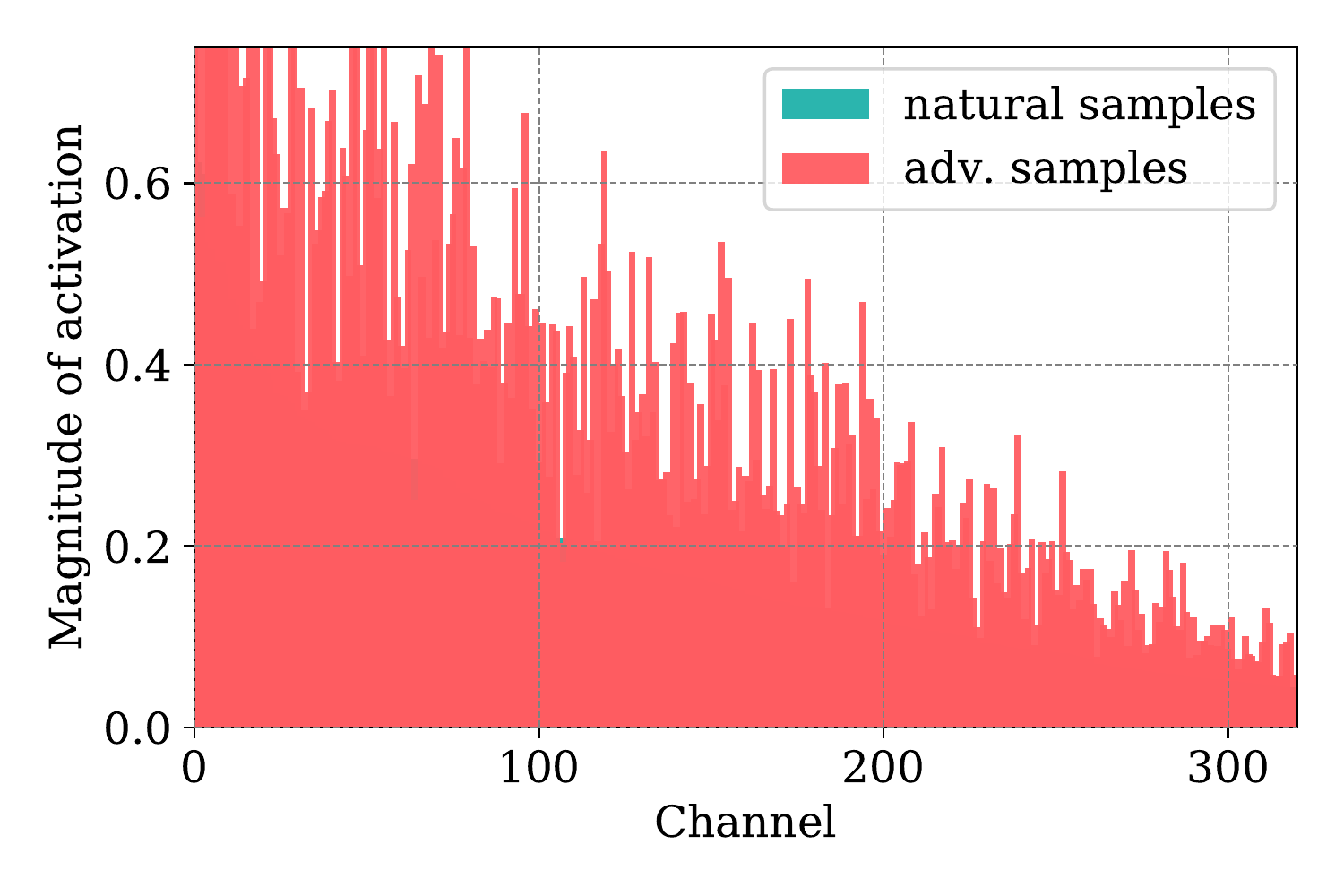}}}%
    \hfill
    \subfloat[\centering TRADES ]{{\includegraphics[width=0.42\linewidth]{ 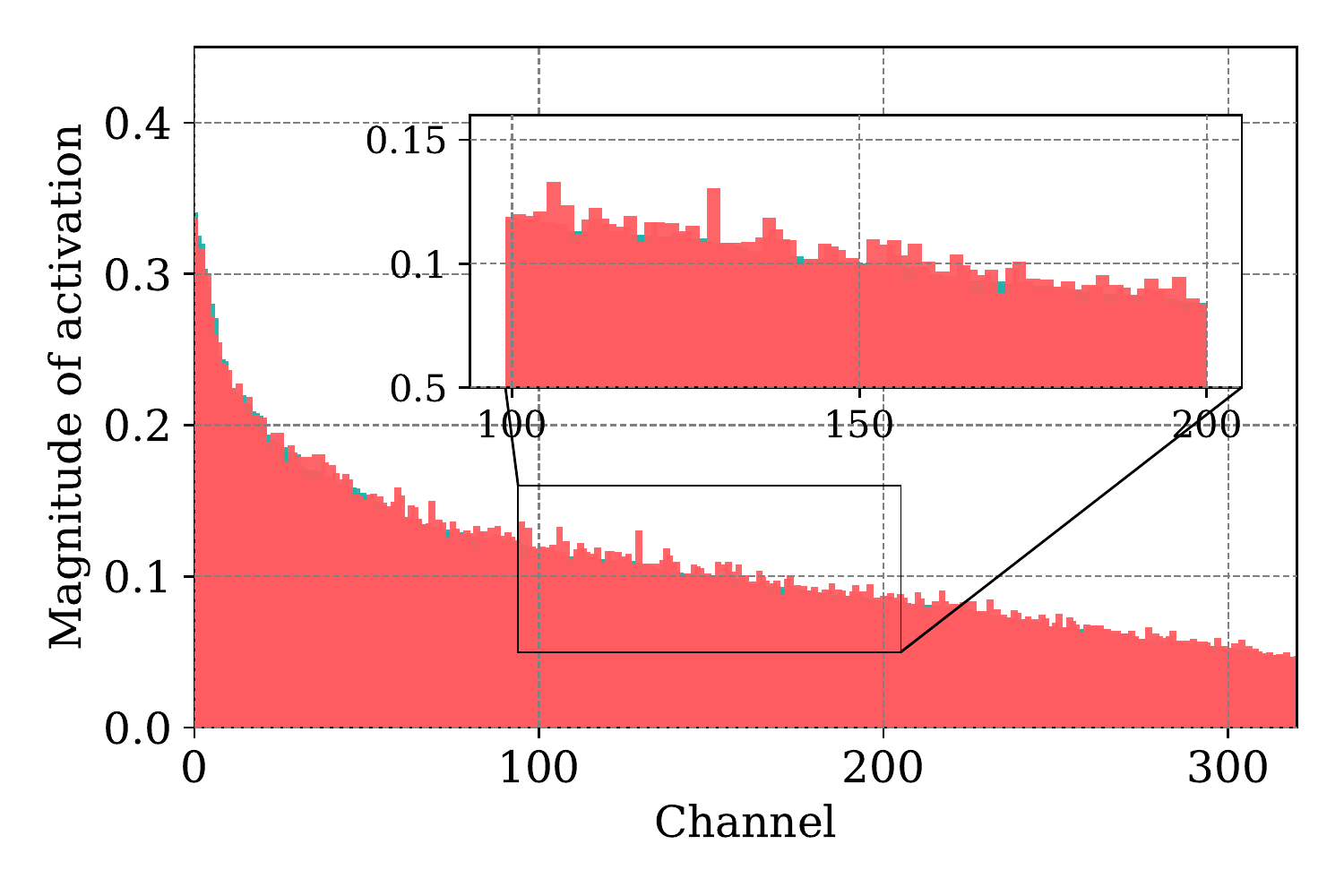}}}%
    \hfill
    \subfloat[\centering ALP]{{\includegraphics[width=0.42\linewidth]{ 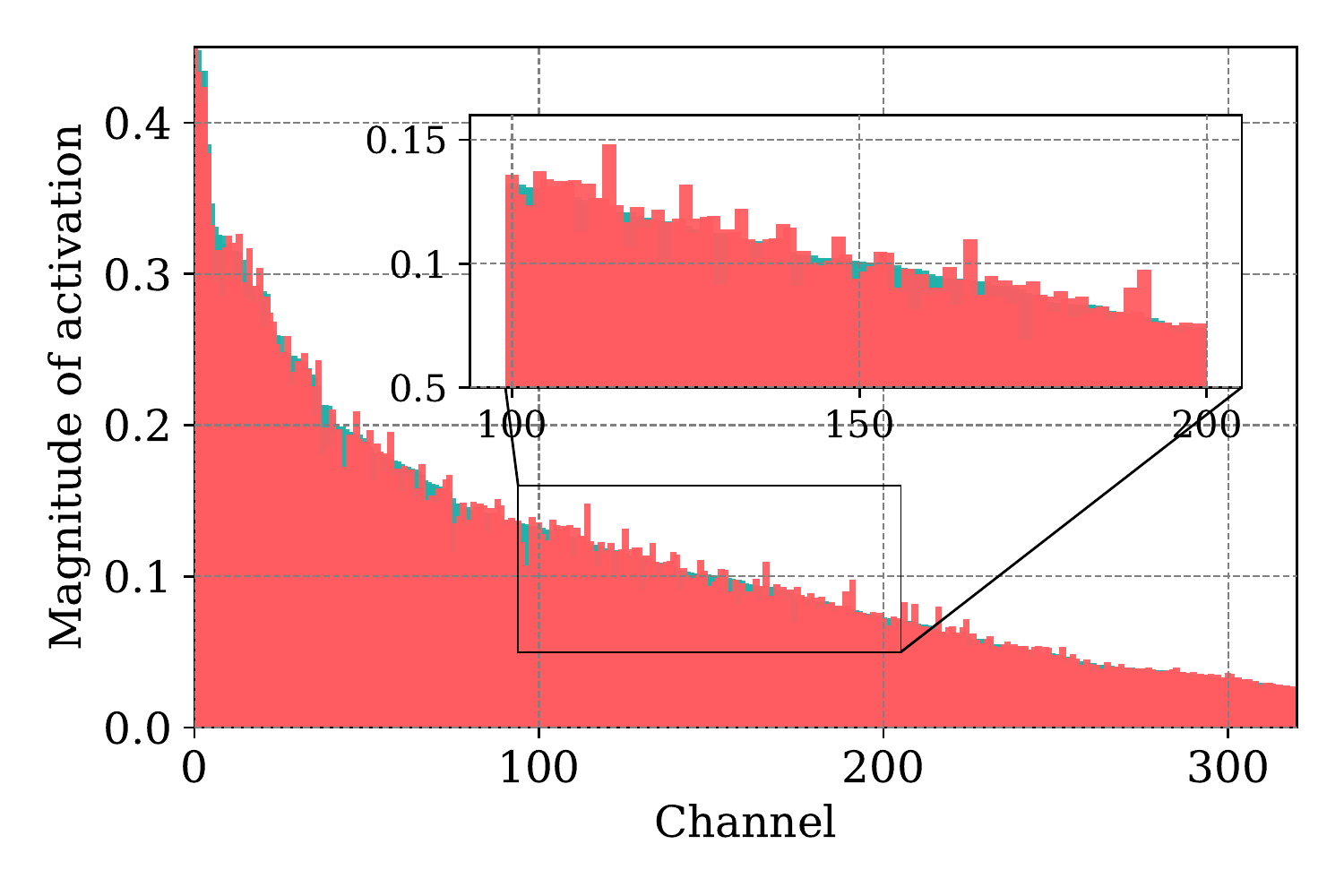}}}%
    \hfill
    \subfloat[\centering PGD-AT]{{\includegraphics[width=0.42\linewidth]{ 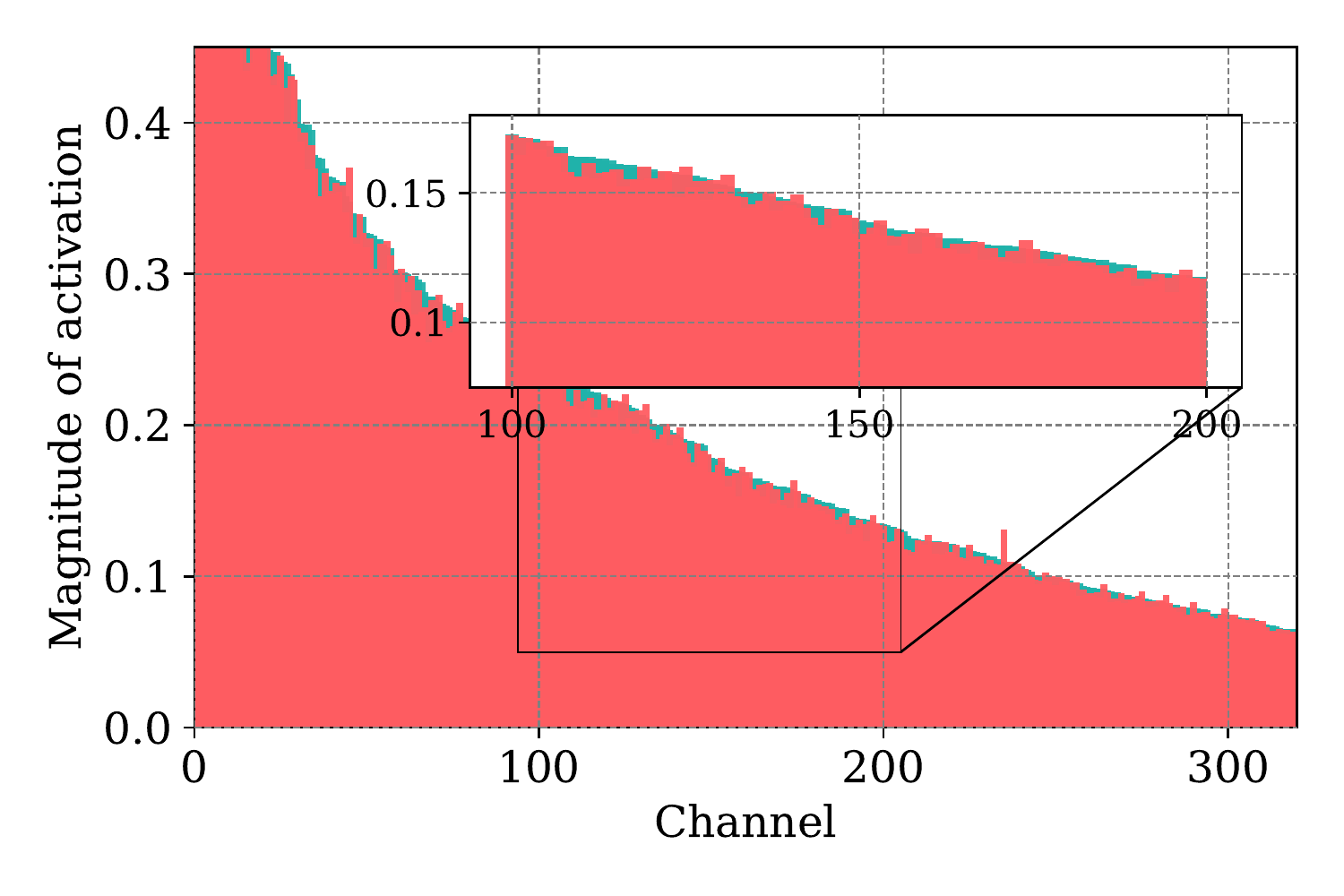}}}%
    \hfill
    \subfloat[\centering SAT  ]{{\includegraphics[width=0.42\linewidth]{ 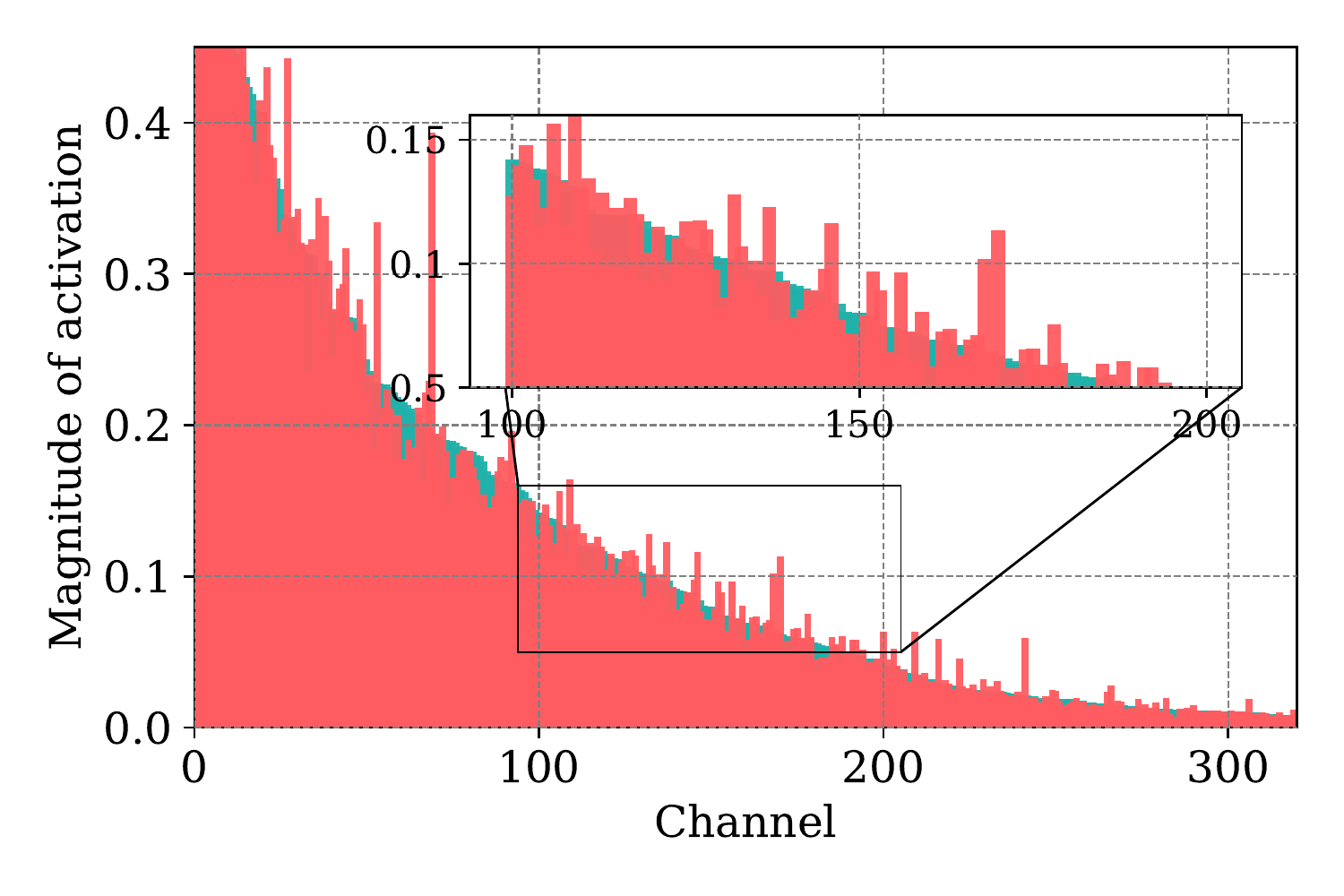} }}%
    \hfill
    \subfloat[\centering~\SystemName~(\textit{Ours}) ]{{\includegraphics[width=0.42\linewidth]{ 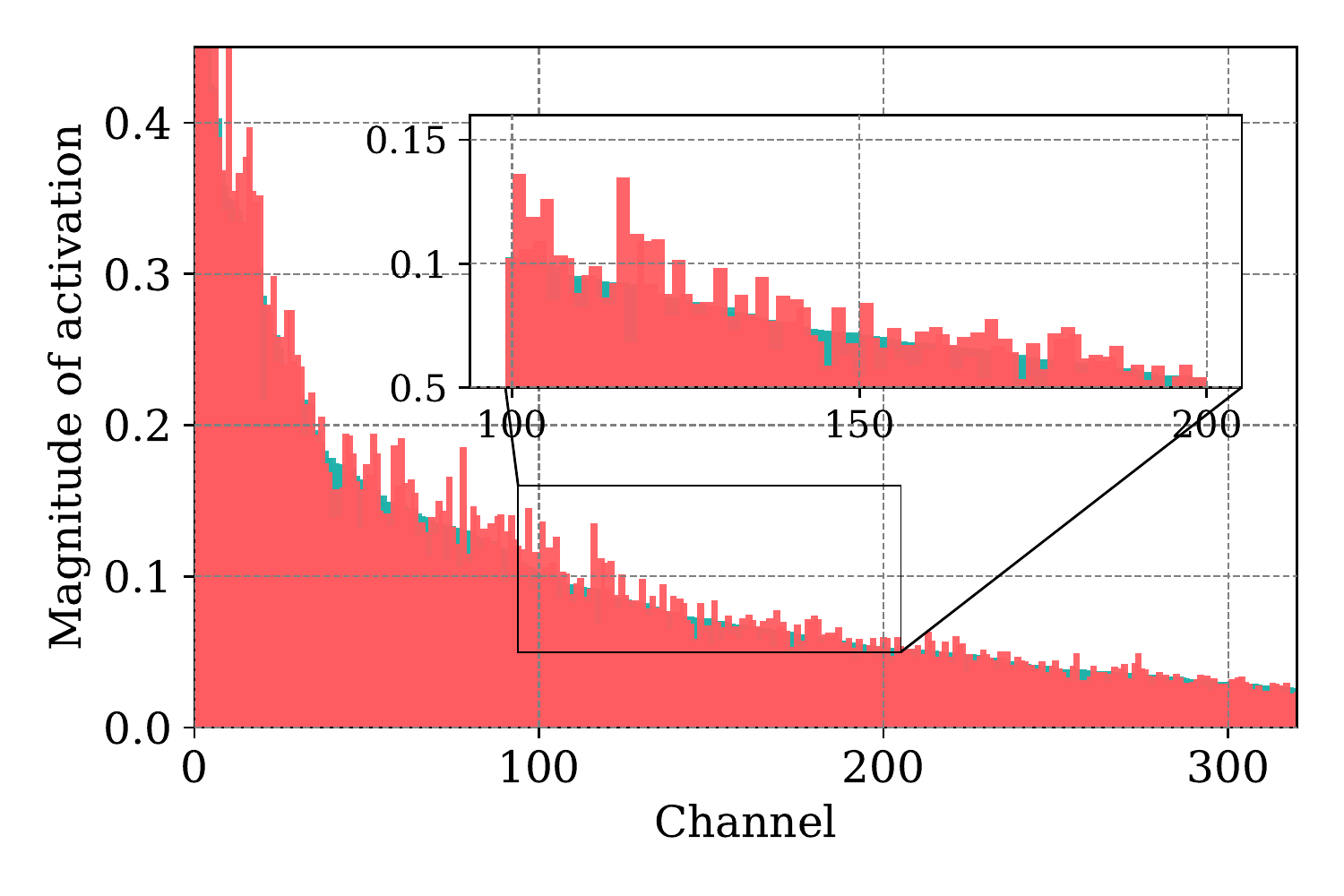}}}%
    \hfill
    \caption{\textbf{Magnitude of activation} at the penultimate layer for models trained with different defense methods. Our \SystemName~ can regulate adversarial samples' magnitudes similar to clean samples' while well suppressing both of them.}%
    \label{fig:mag_activation_all}
\end{figure*}

Figure ~\ref{fig:mag_activation_all} depicts the activation magnitudes at the penultimate layer of WRN34 across various AT frameworks. Although AT methods manage to bring the adversarial magnitudes closer to their clean counterparts, the magnitudes generally remain elevated, with PGD-AT being especially prominent. Through a balanced integration of the input Jacobian matrix and output distributions, our proposed method effectively mitigates the model's susceptibility to perturbed samples. \\

\section{Broader Impact}
Utilizing machine learning models in real-world systems necessitates not only high accuracy but also robustness across diverse environmental scenarios. The central motivation of this study is to devise a training framework that augments the robustness of Deep Neural Network (DNN) models in the face of various adversarial attacks, encompassing both white-box and black-box methodologies. To realize this objective, we introduce the \SystemName~framework, an innovative approach that refines traditional Jacobian regularization techniques and aligns output distributions. This research represents a significant stride in synergizing adversarial training with input-output Jacobian regularization—a combination hitherto underexplored—to {construct} a more resilient model.

\clearpage

\end{document}